\pdfoutput=1

\documentclass[11pt]{article}

\usepackage[]{acl}

\usepackage{times}
\usepackage{latexsym}
\usepackage{amsmath,amssymb,amsfonts}
\usepackage{graphicx}
\usepackage{textcomp}
\usepackage{xcolor}
\usepackage[noend]{algpseudocode}
\usepackage{url}
\usepackage{algorithmicx,algorithm}
\usepackage{pdflscape}
\usepackage{multirow}
\usepackage{makecell}
\usepackage{booktabs}
\usepackage{rotating}
\usepackage{amsmath}
\usepackage{colortbl}
\usepackage{amsfonts}
\usepackage{subfigure}
\usepackage{colortbl}

\usepackage[T1]{fontenc}

\usepackage[utf8]{inputenc}

\usepackage{microtype}

\title{\textsc{ZeroGen}: Zero-shot Multimodal Controllable Text Generation with Multiple Oracles}

\author{Haoqin Tu, Bowen Yang, Xianfeng Zhao\\
State Key Laboratory of Information Security, Institute of Information Engineering,\\
School of Cyber Security, University of Chinese Academy of Sciences\\
\texttt{tuisaac163@gmail.com}, \texttt{\{yangbowen,zhaoxianfeng\}@iie.ac.cn}
}

\begin{document}
\maketitle
\begin{abstract}
Automatically generating textual content with desired attributes is an ambitious task that people have pursued long. Existing works have made a series of progress in incorporating unimodal controls into language models (LMs), whereas how to generate controllable sentences with multimodal signals and high efficiency remains an open question. To tackle the puzzle, we propose a new paradigm of zero-shot controllable text generation with multimodal signals (\textsc{ZeroGen}). Specifically, \textsc{ZeroGen} leverages controls of text and image successively from token-level to sentence-level and maps them into a unified probability space at decoding, which customizes the LM outputs by weighted addition without extra training. To achieve better inter-modal trade-offs, we further introduce an effective dynamic weighting mechanism to regulate all control weights. Moreover, we conduct substantial experiments to probe the relationship of being in-depth or in-width between signals from distinct modalities. Encouraging empirical results on three downstream tasks show that \textsc{ZeroGen} not only outperforms its counterparts on captioning tasks by a large margin but also shows great potential in multimodal news generation with a higher degree of control. Our code will be released at \url{https://github.com/ImKeTT/ZeroGen}.
\end{abstract}

\begin{figure}[!t]
\centering
\includegraphics[width=0.95\linewidth]{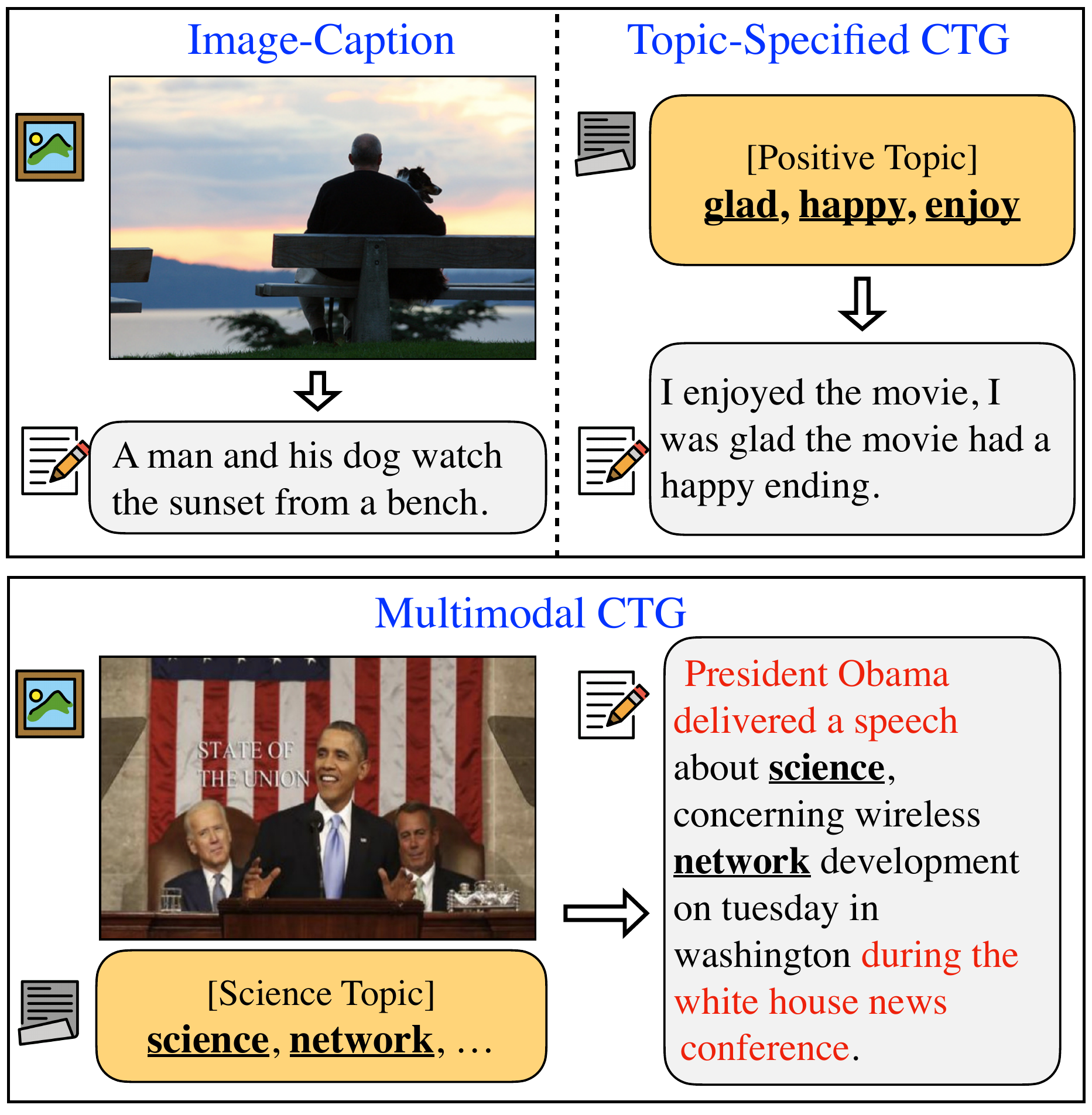}
\caption{Traditional CTG only has unimodal guidance (up), while our \textsc{ZeroGen} follows \textcolor{blue}{Multimodal CTG} (down) that incorporates multimodal controls to generate relevant texts. We mark words/sentences that are relevant to the \underline{\textbf{textual control}} or \textcolor{red}{visual control}.}
\vspace{-0.1in}
\label{fig:demonstration}
\end{figure}

\section{Introduction} \label{sec:intro}
Large-scale pre-trained models (PTMs) have recently achieved great success and become a milestone in the field of AI. Owing to their sophisticated pre-training objectives and huge model parameters, PTMs can benefit a variety of downstream tasks just like \textit{Oracle}s. In the domain of language, pre-trained language models (PLMs) have become a cornerstone of versatile generation tasks including controllable text generation (CTG). By controlling the presence of certain linguistic attributes, these PLMs can be trained to generate texts with desired aspects such as length, and topic \cite{kikuchi2016controlling,ficler2017controlling}. Conventional approaches usually construct a conditional LM with supervision (e.g., by fine-tuning), which is unscalable due to the combinatorially numerous conceivable compositions and the lack of annotated data \cite{keskar2019ctrl,liu2022composable}. Most recent studies have begun to look into ``plug-and-play'' (PnP) solutions. Those techniques plug in arbitrary restrictions to guide the generation of desired sentences with PLMs and little training expenses. And the control signals of this paradigm are typically limited to unimodal domains, such as provided keywords or topics \cite{dathathri2019plug,pascual2021plug,yang2021fudge,tu2022pcae}. Rapidly, the PnP fashion has been adopted to bridge multimodal knowledge, recent works have introduced pre-trained multimodal models like CLIP \cite{radford2021learning} into cross-modal tasks with vision-only controls such as captioning. These approaches obtained exceptional performances with minimal or no task-oriented training \cite{su2022language,tewel2022zerocap,nukrai2022text}.

On the one hand, meaningful interactions between human speakers often necessitate real-world experiences \cite{bisk2020experience}, and the text-only instruction alone may not be sufficient to fulfill such communication purpose \cite{harnad1990symbol}. As a result, using unimodal controls for CTG may conflict between how to reliably regulate current PLMs and real-world scenarios (e.g., multimodal controlled news generation in Figure~\ref{fig:demonstration}). On the other hand, unlike some keyword-guided PnP works \cite{pascual2021plug,gu2022improving}, models that incorporate visual guidance into language generation insert constant controls at LM decoding instead of considering the dynamic nature of such process, which may lead to task under-performance \cite{su2022language,tewel2022zerocap}.


To overcome those shortcomings, we take a step further to extend the current unimodal PnP paradigm into a multimodal setting and propose \textsc{ZeroGen}. To accomplish multimodal CTG task, we are aware that inputs from different domains affect different granularities of presences in texts. As shown in Figure~\ref{fig:demonstration}, while textual control steers generated news to the science topic by presenting related keywords, visual control provides more abundant ambient information by producing sentence descriptions. In order to plug in multimodal signals, we propose to unify controls into the LM output probability using token- or sentence-level similarity with several \textit{Oracle}s. Specifically, we first regard the textual guidance as the token-level similarity between keywords and the LM vocabulary from a textual \textit{Oracle} before decoding, then we incorporate such guidance to LM outputs by weighted addition at generation. For visual guidance, we use a multimodal score \cite{su2022language} based on sentence-level probability determined by a multimodal \textit{Oracle}. Finally, we employ beam search to find the token with the highest score at each step. To adapt to the dynamic nature of LM decoding and further promote model performance, we provide a dynamic weighting mechanism on the word-level that can not only enhance visual information expression but also maintain output fluency. \looseness=-1

We conduct three tasks (image captioning, stylized captioning, and controllable news generation) with \textsc{ZeroGen}. We explore the relationship between textual and visual control being either vertical or lateral. Specifically, in two captioning tasks, textual objects of the image extend the visual signal as a complement (vertical extension). For news generation, a collection of positive or negative words are used to embody generated news a specific sentiment (lateral extension). The effectiveness of our approach in providing better captions and easily controlled news is demonstrated by results on both automatic metrics and human evaluations.

\noindent \textbf{Contributions.} (1) We explore the task of multimodal controllable text generation under zero-shot setting and propose $\textsc{ZeroGen}$ that utilizes token- and sentence-level multimodal guidance to fulfill this task. (2) We present a dynamic weighting scheme on the word-level that can be applied to different modalities and boost the fluency and controllability of generated texts. (3) Extensive experiments on two captioning tasks and the controllable news generation task not only justify the effectiveness of \textsc{ZeroGen} but also investigate the relationship between different types of modal controls.\looseness=-1

\begin{figure*}[!t]
\centering
\includegraphics[width=0.95\linewidth]{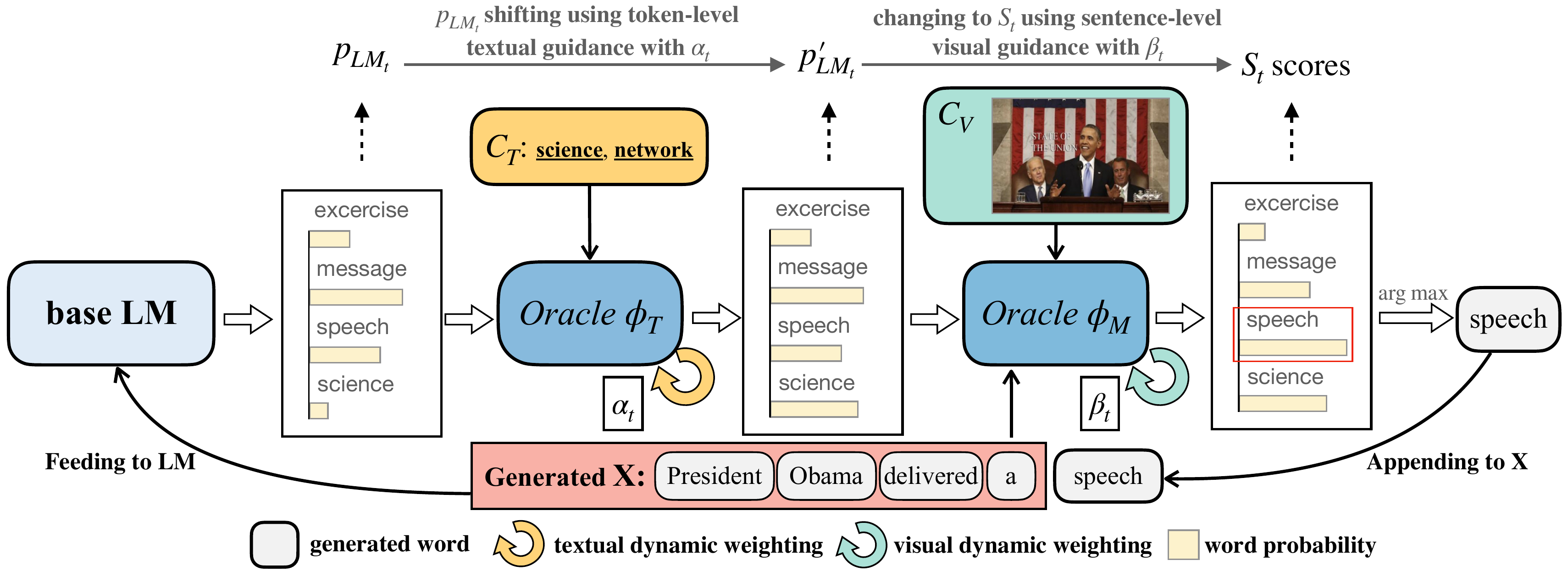}
\caption{Workflow of \textsc{ZeroGen} at decoding step $t$. Through multiple LM output changing stages, \textsc{ZeroGen} is essentially a decoding scheme that finds a word related to both textual ($\textbf{C}_T$) and visual control ($\textbf{C}_V$) at each step. It then feeds the word back to the base LM for the future conditional generation.}
\vspace{-0.1in}
\label{fig:plot}
\end{figure*}

\section{Related Work}
\paragraph{Efficient Image Captioning.} The prerequisite of supervised captioning for a large amount of paired image-text data is unrealistic in real-life scenarios. Various attempts have been made to reduce the dependence on large paired image-text data. For example, some works \cite{anderson2018partially,laina2019towards,chen2020simple,honda2021removing} have sought to incorporate objects from given images into model training. Despite their efficiency in comparing supervised methods, they still need to be trained with partial cross-modal guidance as supervision. CLIP \cite{radford2021learning} as a milestone for vision-language alignment has shown impressive zero-shot capabilities on various multimodal generation tasks. For example, \citet{tewel2022zerocap} proposed the first zero-shot captioning model with CLIP and a base LM (i.e., GPT-2). It constantly updates the model's transformer cache under the direction of CLIP guidance decoding. Nevertheless, it still demands gradient computation and optimization during generation, introducing additional generation overhead. \citet{su2022language} proposed \textsc{MAGIC} that utilizes a token decoding score based on CLIP to produce plausible captions without task-specified training. Most recently, \citet{nukrai2022text} employs text-only training with gaussian noises parameterized by a few images to connect CLIP and the base LM textual embedding. Still, \citet{nukrai2022text} requires a small amount of external visual knowledge during training. As for our model, \textsc{ZeroGen} expands \textsc{MAGIC} with additional capabilities to facilitate multimodal guided generation with dynamic weighting, supporting several downstream applications while keeping its ability to transfer to different base LMs. Most recently, \cite{zeng2023conzic} proposed to employ sample-based sequential polishing during language decoding to produce plausible and fluent captions.

\paragraph{PnP Controllable Text Generation.} To avoid excessive training costs from fine-tuning PLMs into CTG tasks, researchers have turned their attention to specialized training-free methods such as the ``plug-and-play'' (PnP) framework by \citet{dathathri2019plug}. This framework can be used along an existing generative LM (the base LM) with minimum or no training procedure between PnP components and the base LM. In comparison to conventional methods, these PnP approaches typically follow two aspects. \textit{In-model guidance} approaches including ``prompt tuning'' \cite{lester2021power} that either aim at optimizing the input prompts and additional parameters that are fed into the base LM \cite{houlsby2019parameter,li2021prefix,lester2021power} or seek to alter certain hidden representations that are not model input or output layers, by plugging a trainable model into the middle of the base LM \cite{dathathri2019plug,duan2020pre,mai2020plug,tu2022pcae}. \textit{Out-model guidance} techniques, on the contrary, focus on building controllable language models that only modify the output probabilities from the base LMs at inference time \cite{krause2021gedi,pascual2021plug,yang2021fudge,liu2021dexperts,krause2021gedi}. And our \textsc{ZeroGen} belongs to the last category that only imposes control signals at LM decoding.

\section{\textsc{ZeroGen} Methodology}
For the multimodal CTG task, we formally define it as: given the visual control $\textbf{C}_V$ (i.e., an image) and $N$ representative words from a topic or an image as the textual control $\textbf{C}_T = \{C_{T_1}, ..., C_{T_N}\}$, we aim at getting the textual output $\textbf{X}=\{x_{1}, x_{2}, ...\}$ to meet the two control aspects simultaneously. \looseness=-1

\textsc{ZeroGen} focuses on the output probability space of the base LM. As shown in Figure~\ref{fig:plot}, at decoding step $t$, it first adjusts the original LM output probability $p_{\text{LM}_t}$ to $p'_{\text{LM}_t}$ follows the token-level textual guidance from keywords-vocabulary similarities, then it completes word searching on $p'_{\text{LM}_t}$ using a sentence-level multimodal scoring function and beam search. Note that, instead of calculating the token similarity constantly \cite{pascual2021plug}, we only compute it once before decoding and turning it into the overall textual control with options. Finally, being processed on word-level, the dynamic weighting scheme is applied to regulate both control weights for every generation step. \looseness=-1


\subsection{Token-level Textual Guidance} \label{sec:textual-control}
Since the appearance of keywords from a certain topic can drive sentences in such direction, we consider token-level similarity between LM tokens and keywords in $\textbf{C}_{T}$ as the textual guidance. To avoid the additional computational costs, we unify the textual control into probability space by a set of cosine similarities between word $C_{T_n}\in \textbf{C}_{T}$ and the full base LM vocabulary $\textbf{V}\in \mathbb{R}^{V}$ before decoding. These word similarities are obtained using the textual \textit{Oracle} $\phi_T$ (e.g., pre-trained word embedding):
\begin{equation} \nonumber
\begin{aligned}
& p(\textbf{V}, \textbf{C}_{T}) = \left\{\cos(\phi_T(\textbf{V}), \phi_T(C_{T_n}))\right\}_{n=1}^N,
\end{aligned}
\end{equation}
where $p(\textbf{V}, \textbf{C}_{T})\in \mathbb{R}^{N\times V}$, $V$ is the vocabulary size. To fully utilize all the given keywords, we explore three selection methods at time $t$ when $N>1$ to get the overall textual control $p_t(\textbf{C}_T)\in \mathbb{R}^V$:

\paragraph{Step-wise Random (SR):} we first provide changing controls through the generation. At different steps, we sample one keyword-vocabulary similarity uniformly from $p(\textbf{V}, \textbf{C}_{T})$ as textual guidance.
\paragraph{Mean Pooling (MP):} an intuitive way to consider all textual information is to average their guiding similarities w.r.t. $\textbf{V}$ across distinct keywords.
\paragraph{Word-wise Max (WM):} for every token $w$ from $\textbf{V}$, we choose the most similar keyword in $\textbf{C}_{T}$ with $w$ (with the highest cosine similarity score) to compute its guiding probability and compose all the highest similarities together as $p_t(\textbf{C}_T)$.


The overall textual control $p_t(\textbf{C}_T)\in \mathbb{R}^V$ is available after this selection, we introduce it to $p_{\text{LM}_t}$ as a control bias by simple addition operation with weighting: $p'_{\text{LM}_t} = p_{\text{LM}_t} + \alpha \times p_t(\textbf{C}_T)$.
\subsection{Sentence-level Visual Guidance} \label{sec:visual_control_method}
Image information carries more general and higher-level global information than a single word. As discussed in Sec.~\ref{sec:intro}, we thus consider sentence-level similarity between generated texts and visual control $\textbf{C}_V$ as the visual guidance.

We employ scoring function $S_t$ for word $w\in \textbf{V}$ at $t$-th step with weighted visual guidance as in \citet{su2022language}, and use beam search for generation:
\begin{equation} \nonumber
    \begin{aligned}
    & S_{t}\left(w, \textbf{C}_V\mid x_{<t}, W_t^{(k)}\right)=\\&
    \left\{\begin{aligned}
         &\begin{aligned} p'_{\text{LM}_t}&(w\mid x_{<t}) + \\& \beta \times \frac{e^{p_{\phi_M}\left(\left[x_{<t};w\right], \textbf{C}_V\right)}}{\sum_{z\in W_t^{(k)}} e^{p_{\phi_M}\left(\left[x_{<t};z\right], \textbf{C}_V\right)}} \end{aligned}, \text{if } w\in W_t^{(k)} \\
         & -\inf,  \qquad\qquad\qquad\qquad\qquad\quad\quad \text{otherwise}.
    \end{aligned}\right.
    \end{aligned}
\end{equation}
Here $[x_{<t};w]$ means appending $w$ to the generated texts before step $t$, $W_t^{(k)}$ is the searching beam consists of words with the $k$ highest probabilities in $p'_{\text{LM}_t}$. In detail, we bridge texts and $\textbf{C}_V$ using multimodal \textit{Oracle} $\phi_M$ (e.g., CLIP) and compute their similarity: $p_{\phi_M}([x_{<t};w], \textbf{C}_V) = \cos (\phi_M([x_{<t};w]), \phi_M(\textbf{C}_V))$. Our final goal is to find $x_t = \arg\max_{w\in \textbf{V}} S_{t}(w, \textbf{C}_V\mid x_{<t}, W_t^{(k)})$.\looseness=-1

\subsection{Multimodal Dynamic Weighting}
To further boost the model performance and make the model get attuned to different generation steps, a novel dynamic weighting mechanism is proposed to achieve step-wise multimodal weights adjustment. Concretely, we replace $\alpha, \beta$ with dynamic $\alpha_t, \beta_t$ severally. The design should consider such principles: (1) It is necessary to seek a certain balance between textual control (i.e., shifts the LM output probability) and the original LM modeling to avoid inconsistent outputs. (2) During generation, visual-relevant words ought to be encouraged, while those irrelevant are punished. Since the smallest comprehensible output of an LM is one word, we then apply this framework on the word-level.

\paragraph{Dynamic $\boldsymbol{\alpha_t}$.} To maintain the original language modeling ability, and also make the most out of provided textual guidance. We re-scale the textual control using a step-wise weighting calibration that incorporates the original LM output confidence $p_{\text{LM}_t}$. Specifically, we compute the average probability of $\hat{N} \in [1, N]$ keywords in $\textbf{C}_T$ from the unshifted LM output as the $t$-th textual control weight:
\begin{equation} \nonumber
    \begin{aligned}
    & D_T = \sum_{n=1}^{\hat{N}}\frac{p_{\text{LM}_t}\left(C_{T_n}\mid x_{<t}\right)}{\hat{N}},
    & \alpha_t = \min \left( \frac{D_T}{\lambda}, \hat{\alpha}\right).
    \end{aligned}
\end{equation}
If $D_T$ is high, keywords from $\textbf{C}_T$ are encouraged to be spoken. Since this is the exact time when unchanged base LM has high confidence to produce these words, we can avoid jeopardizing output fluency while generating controlled texts.

\paragraph{Dynamic $\boldsymbol{\beta_t}$.} To reward the generation stages where all words in $W_t^{(k)}$ are highly associated with the knowledge in $\textbf{C}_V$ and penalize those do not, we employ average word-level similarity between current candidate words and the visual control:
\begin{equation} \nonumber
    \begin{aligned}
    & D_V = \sum_{w\in W_t^{(k)}}\frac{p\left(w, \textbf{C}_V\right)}{k},
    & \beta_t = \min \left( \frac{D_V}{\lambda}, \hat{\beta}\right).
    \end{aligned}
\end{equation}
If $D_V$ is high, words in $W_t^{(k)}$ are relevant to $\textbf{C}_V$ and should be expressed with a higher chance.

Inspired by \citet{gu2022improving}, $\lambda$ in this framework serves as a threshold that amplifies the control signal if $D_V$ or $D_T$ is larger than it and vice versa. Meanwhile, $\hat{\alpha}, \hat{\beta}$ are weighting upper bounds.

\begin{table*}
\setlength\tabcolsep{2.3pt}
\centering
\small
\begin{tabular}{l|cccccc|ccccccc} 
\toprule[1pt]
\multirow{2}{*}{Model}      & \multicolumn{6}{c|}{\texttt{MS-COCO}}                                                                                   & \multicolumn{6}{c}{\texttt{Flickr30k}}                                                                          & \multirow{2}{*}{Speed}  \\ 
\cmidrule{2-13}
                            & B@1 $\uparrow$ & B@4 $\uparrow$ & M $\uparrow$  & R-L $\uparrow$ & CIDEr $\uparrow$ & SPICE $\uparrow$         & B@1 $\uparrow$ & B@4 $\uparrow$ & M $\uparrow$  & R-L $\uparrow$ & CIDEr $\uparrow$ & SPICE $\uparrow$ &                         \\ 
\midrule
\multicolumn{14}{c}{\textit{Weakly Supervised Approaches}}                                                                                                                                                                                                                              \\
\midrule
\textsc{IC-SME} \shortcite{laina2019towards} & -           & 6.5            & 12.9          & 35.1           & 22.7             & {-}  & -              & 7.9              & 13.0             & 32.8              & 9.9                & -                & -                       \\
\textsc{S2S-GCC} \shortcite{honda2021removing} & 50.4           & 7.6            & 13.5          & 37.3           & 31.8             & {8.4}  & -              & -              & -             & -              & -                & -                & -                       \\
\textsc{CapDec} \shortcite{nukrai2022text}  & 69.2           & 26.4           & 25.1          & 51.9           & 91.8             & {-}    & 55.5           & 17.7           & 20.0          & 43.9           & 39.1             & -                & -                       \\
\midrule
\multicolumn{14}{c}{\textit{Unsupervised Approaches}}                                                                                                                                                                                                       \\
\midrule
\textsc{CLIPRe}                      & 39.5           & 4.9            & 11.4          & 29.0           & 13.6             & 5.3                      & 38.5           & 5.2            & 11.6          & 27.6           & 10.0             & 5.7              & -                       \\
\textsc{ZeroCap} \shortcite{tewel2022zerocap}                     & 49.8           & 7.0           & 15.4          & 31.8           & 34.5           & 9.2                     & 44.7           & 5.4            & 11.8          & 27.3           & 16.8             & 6.2              & 1.0 $\times$            \\ 
\textsc{MAGIC} \shortcite{su2022language}                       & 56.5           & 12.4           & 17.3          & 39.6           & 48.3             & 11.2                     & 43.3           & 6.8            & 12.3          & 30.8           & 20.5             & 6.8              & \textbf{26.6} $\times$           \\ 
\textsc{ConZIC} \shortcite{zeng2023conzic}                       & -           & 1.3           & 11.5          & -           & 12.8             & 5.2                     & -           & -            & -          & -           & -             & -              & -            \\
\textsc{DeCap} \shortcite{li2023decap}                       & -           & 8.9           & 17.5          & -           & 50.6             & \textbf{13.1}                     & -           & -            & -          & -           & -             & -              & -            \\
\midrule
\textsc{ZeroGen}              & \textbf{59.4}  & \textbf{15.5}  & \textbf{18.7} & \textbf{42.3}  & \textbf{55.4}    & 12.1         & \textbf{54.9}  & \textbf{13.1}  & \textbf{15.2} & \textbf{37.4}  & \textbf{26.4}    & \textbf{8.3}     & 16.4 $\times$           \\
\quad -\texttt{TDW}            & 58.9           & 15.2           & 18.4          & 41.8           & 54.4             & 11.9                     & 54.1           & 12.8           & 14.7          & 36.8           & 24.5             & 7.7              & 16.5 $\times$           \\
\quad -\texttt{T}     & 58.6           & 14.7           & 17.4          & 41.3           & 51.7             & 11.8                     & 53.3           & 11.9           & 14.3          & 36.2           & 24.1             & 7.5              & 18.6 $\times$           \\
\quad -\texttt{VDW}            & 57.0           & 12.6           & 17.6          & 39.7           & 49.7             & 11.6                     & 49.2           & 6.4            & 14.1          & 32.4           & 22.9             & 7.7              & 22.5 $\times$           \\
\quad -\texttt{DW}                   & 57.0           & 12.6           & 17.6          & 39.7           & 49.7             & 11.6                     & 47.7           & 7.1            & 13.8          & 32.3           & 21.9             & 7.6              & 21.6 $\times$           \\
\bottomrule[1pt]
\end{tabular}
\caption{Captioning results of \textsc{ZeroGen} with only 1 object as $\textbf{C}_T$ (i.e., $N=1$) on \texttt{MS-COCO} and \texttt{Flickr30k}. \textsc{ZeroGen} outperforms most baselines with tolerable efficiency sacrifice. \texttt{T}, \texttt{TDW}/\texttt{VDW}, \texttt{DW} represent textual control, textual/visual dynamic weighting and two dynamic weighting schemes combined respectively.}
\label{tab:caption_main}
\end{table*}

\begin{table}[!t]
\setlength\tabcolsep{2pt}
\centering
\small
\begin{tabular}{l|ccccc} 
\toprule[1pt]
{Model}      & B@1 $\uparrow$ & B@4 $\uparrow$ & M $\uparrow$  & R-L $\uparrow$ & CIDEr $\uparrow$                          \\ 
\midrule
\textsc{ZeroGen} ($N=1$)              & 59.4  & 15.5  & \textbf{18.7} & 42.3  & 55.4   \\
\textsc{ZeroGen} ($N=2$)              & 60.1  & 15.6  & 18.5 & 42.3  & 55.9    \\
\textsc{ZeroGen} ($N=3$)              & 60.4  & 15.6  & 18.6 & 42.3  & 56.5\\
\textsc{ZeroGen} ($N=4$)              & 60.5  & 15.7  & \textbf{18.7} & \textbf{42.4}  & 57.0   \\
\textsc{ZeroGen} ($N=5$)              & \textbf{60.6}  & \textbf{15.8}  & \textbf{18.7} & \textbf{42.4}  & \textbf{57.1} \\
\bottomrule[1pt]
\end{tabular}
\caption{Captioning results of \textsc{ZeroGen} on \texttt{MS-COCO} with varied size $N$ for textual control $\textbf{C}_T$.}
\label{tab:n_obj}
\end{table}
\begin{table}[!t]
\setlength\tabcolsep{2.5pt}
\centering
\small
\begin{tabular}{l|ccccc} 
\toprule[1pt]
{Model}      & B@1 $\uparrow$ & B@4 $\uparrow$ & M $\uparrow$  & R-L $\uparrow$ & CIDEr $\uparrow$                          \\ 
\midrule
\textsc{ZeroGen} \textbf{SR}   & 59.6  & 15.3  & 18.4 & 42.1  & 55.5 \\
\textsc{ZeroGen} \textbf{MP}       & 59.9  & 15.2  & 18.3 & 42.0  & 55.2\\
\textsc{ZeroGen} \textbf{WM}      & \textbf{60.6}  & \textbf{15.8}  & \textbf{18.7} & \textbf{42.4}  & \textbf{57.1}\\

\bottomrule[1pt]
\end{tabular}
\caption{Captioning results of \textsc{ZeroGen} with $N=5$ for $\textbf{C}_T$ on \texttt{MS-COCO} with three $p_t(\textbf{C}_T)$ options.}
\label{tab:pc_selection}
\end{table}
\section{General Implementations and Baselines}
\paragraph{General Implementations.} We take SimCTG \cite{su2022contrastive} as our base LM and first fine-tune it on every dataset with text-only data like previous works. Since \textsc{ZeroGen} follows the zero-shot paradigm, it can leverage any off-the-shelf LM and empower it a pair of eyes. For \textit{Oracle}s, we employ GloVe \cite{pennington2014glove} as the textual \textit{Oracle} $\phi_T$ and CLIP \cite{radford2021learning} as the multimodal \textit{Oracle} $\phi_M$. The $\hat{N}$ for $\alpha_t$ is $N$ itself on two captioning tasks, while $\hat{N}=2$ on controllable news generation task through ablation study. The amplifying factor $\lambda$ is $0.2$ throughout the paper. See Appendix~\ref{app:details} for full model details.

\paragraph{Baseline Models.} For the image captioning task, we select both weakly supervised and unsupervised methods as our baselines, (1) \textsc{IC-SME} \cite{laina2019towards}, \textsc{S2S-GCC} \cite{honda2021removing}, and \textsc{CapDec} \cite{nukrai2022text} are three weakly supervised approaches, the former two adapt neural network modules to align visual features with pseudo captions, \textsc{CapDec} introduces CLIP guidance and few images in training. (2) \textsc{CLIPRe}, \textsc{ZeroCap} \cite{tewel2022zerocap}, and \textsc{MAGIC} \cite{su2022language} are three zero-shot methods, which follow the retrieval manner, CLIP-guided gradient update, and decoding scheme respectively. For fair comparisons, we use the same base LM as ours for \textsc{ZeroCap} and \textsc{MAGIC}. In stylized captioning, \textsc{MemCap} \cite{zhao2020memcap} is additionally considered.\looseness=-1


\begin{table*}[!t]
\setlength\tabcolsep{2.8pt}
\centering
\small
\begin{tabular}{l|cccccc|cccccc} 
\toprule[1pt]
\multirow{2}{*}{Model} & \multicolumn{6}{c|}{\texttt{FlickrStyle10k~Romantic}}           & \multicolumn{6}{c}{\texttt{FlickrStyle10k~Humorous}}              \\ 
\cmidrule{2-13}
                       & B@1 $\uparrow$ & B@3 $\uparrow$ & M $\uparrow$  & R-L $\uparrow$ & CIDEr $\uparrow$ & SPICE $\uparrow$ & B@1 $\uparrow$  & B@3 $\uparrow$ & M $\uparrow$   & R-L $\uparrow$ & CIDEr $\uparrow$ & SPICE $\uparrow$ \\ 
\midrule
\textsc{MemCap} \shortcite{zhao2020memcap}                 & 21.2 & 4.8 & 8.4 & -    & 22.4  & -     & 19.9 & 4.3 & 7.4  & -    & 19.4  & -      \\
\textsc{ZeroCap} \shortcite{tewel2022zerocap}                & 19.3 & 2.7 & 7.6 & 16.5 & 14.9  & 7.0   & 18.4 & 2.7 & 7.7  & 16.5 & 15.6  & 7.7    \\
\textsc{MAGIC} \shortcite{su2022language}                  & 23.3 & 4.9 & 8.6 & 21.7 & 24.4  & 8.6   & 23.7 & 5.2 & 9.0  & 21.2 & 27.8  & 10.1   \\ 
\textsc{CapDec}$^*$ \shortcite{nukrai2022text}                 & 21.4 & 5.0 & \textbf{9.6} & -    & 26.9  & -     & \textbf{24.9} & 4.3 & \textbf{10.2} & -    & \textbf{34.1}  & -      \\ 
\textsc{ConZIC} \shortcite{zeng2023conzic} & - & 1.2 & 6.1 & - & -  & -   & - & 1.2 & 6.1  & - & -  & -   \\
\midrule
\textsc{ZeroGen}           & \textbf{24.4} & \textbf{5.5} & 9.2 & \textbf{22.3} & \textbf{27.3}  & \textbf{9.8}   & 24.2 & \textbf{5.6} & 9.6  & \textbf{22.0} & 30.5  & \textbf{11.2}   \\
\quad -\texttt{TDW}         & 23.5 & 5.4 & 8.7 & 21.9 & 26.1  & 9.0   & 24.2 & \textbf{5.6} & 9.6  & 21.9 & 30.5  & \textbf{11.2}   \\
\quad -\texttt{T} & 23.3 & 4.9 & 8.6 & 21.8 & 24.7  & 8.6   & 23.7 & 5.2 & 9.1  & 21.2 & 28.3  & 10.2   \\
\quad -\texttt{VDW}         & 24.0 & 5.5 & 9.0 & 22.1 & 26.9  & 9.4   & 24.1 & 5.6 & 9.5  & 21.9 & 30.2  & 11.1   \\
\quad -\texttt{DW}                & 23.4 & 5.1 & 8.7 & 21.8 & 25.0  & 9.0   & 23.8 & 5.3 & 9.1  & 21.4 & 29.1  & 10.3   \\
\bottomrule[1pt]
\end{tabular}
\caption{Stylized captioning results on two subsets of \texttt{FlickrStyle10k} with $N=1$. * meas \textsc{CapDec} is a weakly supervised method that requires additional visual knowledge from several images during training.}
\label{tab:style_caption}
\end{table*}

For controllable news generation task, we take \textsc{MAGIC} and \textsc{MAGIC+PPLM} as two baseline models. Specifically, \textsc{MAGIC+PPLM} is the combination of two existing PnP works that take image and keywords as input respectively. \textsc{PPLM} \cite{dathathri2019plug} is the first controllable PnP LM that requires gradient descents of model hidden states at decoding time. More details and code links of baselines are available in Appendix~\ref{app:baselines}.

\section{Experiments and Analysis}
\subsection{Image Captioning} \label{sec:image_caption}
\paragraph{Dataset and Metrics.} We conduct experiments on \texttt{MS-COCO} and \texttt{Flickr30k} using Karpathy split \cite{karpathy2015deep}. For the visual control, we take images for captioning task as $\textbf{C}_V$. For the textual control, we take textual objects of the corresponding image as $\textbf{C}_T$.\footnote{Textual objects are extracted from each picture ahead of the generation using a pre-trained \textsc{DETR} \cite{carion2020end}.} We use five relevance-based metrics for evaluation: BLEU-1 (B@1), BLEU-4 (B@4) \cite{papineni2002bleu}, METEOR (M) \cite{denkowski2014meteor}, ROUGE-L (R-L) \cite{lin2004automatic}, CIDEr \cite{vedantam2015cider}, and SPICE \cite{anderson2016spice}. Besides, we also compare the decoding speed of \textsc{ZeroGen} against baselines.\footnote{We measure the model's decoding speed on the same machine with one NVIDIA GeForce 3090 GPU sequentially.}

\paragraph{Main Results.} Since both modal controls aim to enhance the model's ability to understand image content and to generate better captions, we consider $\textbf{C}_T$ to be a vertical augmentation (or a complement) of $\textbf{C}_V$ in this task. From results in Table~\ref{tab:caption_main}, we can draw the following conclusions, (1) the proposed \textsc{ZeroGen} model consistently outperforms unsupervised baselines and most of the weakly supervised methods (except \textsc{CapDec}) by a great margin, demonstrating the superiority of the proposed method. (2) Textual guidance as a vertical augmentation of the visual guidance, provides extra information about an image, thus promoting the model performance by more than 2 absolute points in CIDEr on both datasets (comparing model -\texttt{T}). (3) Both dynamic weighting techniques help strengthen the model's capacity, especially \texttt{VDW}, we ascribe this situation to its direct optimization of certain token appearances that are recognized in the image. (4) However, \textsc{ZeroGen} falls short in efficiency comparing \textsc{MAGIC}, but still largely outperforms \textsc{ZeroCap}. This is because additional computations are required for multimodal controls and dynamic weightings, but there is no need for gradient calculation in our model like \textsc{ZeroCap}.

We also make a series of cross-domain evaluations in Appendix~\ref{app:cross_domain_cap}, which further verifies the robustness of \textsc{ZeroGen} across various domains.

\paragraph{Number of Objects in $\boldsymbol{\textbf{C}_{T}}$.} \textit{Is the more objects the better?} To answer this question, we conduct an experiment over \texttt{MS-COCO} with varied numbers of objects from the image (size $N$ in $\textbf{C}_T$) using word-wise max (\textbf{WM}) for $p(\textbf{C}_{T})$ selection. In Table~\ref{tab:n_obj}, we can observe that, as the increase of object number, our model generally performs better on most metrics, which verifies that more textual object guidance brings more information for captioning task. Similar results also prove the answer on \texttt{Flickr30k} as shown in Appendix~\ref{app:BoOct}.

\paragraph{$\boldsymbol{p(\textbf{C}_{T})}$ Selection Method.} In Table~\ref{tab:pc_selection}, the most effective method for $p(\textbf{C}_{T})$ selection is word-wise max (\textbf{WM}). It is attributed to \textbf{WM}'s highlight of textual objects with all their relevant tokens in the vocabulary. While mean pooling (\textbf{MP}) also takes all given keywords into consideration, by presenting them equally, it may introduce biases in token similarity calculation and output controlling. Hence, we use \textbf{WM} for the rest experiments.

\begin{table*}[!t]
\setlength\tabcolsep{5pt}
\centering
\small
\renewcommand\arraystretch{1}
\begin{tabular}{l|ccccc|cccccc} 
\toprule[1pt]
\multicolumn{1}{c|}{\multirow{2}{*}{Model}}          & \multicolumn{5}{c|}{Positive}                 & \multicolumn{5}{c}{Negative}       & \multirow{2}{*}{Speed $\uparrow$}           \\
\cmidrule{2-11}
               & D-2 $\uparrow$  & D-4 $\uparrow$  & C-S $\uparrow$   & $\Delta \text{Acc}$ $\uparrow$    & PPL $\downarrow$   & D-2 $\uparrow$ & D-4 $\uparrow$  &  C-S $\uparrow$   & $\Delta \text{Acc}$ $\uparrow$    & PPL $\downarrow$    \\
\midrule

\textit{Human}$^*$   & 96.25 & 96.98 & 23.36 & {0.00}     & 14.59 & 96.25 & 96.98 & 23.36 & {0.00}     & 14.59 & -      \\
\midrule
$\textsc{MAGIC}^*$ \shortcite{su2022language}          & \textbf{95.62} & \textbf{95.92} & 20.07 & \cellcolor[rgb]{0.929,0.929,0.929}{-}     & 10.01 & \textbf{95.62} & \textbf{95.92} & 20.07 & \cellcolor[rgb]{0.929,0.929,0.929}{-}     & 10.01   & 25.0 $\times$     \\
\quad +\textsc{PPLM} \shortcite{dathathri2019plug}         & 74.22 & 81.44 & 20.44 & \cellcolor[rgb]{0.929,0.929,0.929}{11.00}     & 29.07 & 74.47 & 83.66  & 20.79 & \cellcolor[rgb]{0.929,0.929,0.929}{18.76}     & 27.32    & 1.0 $\times$    \\
\midrule
\textsc{ZeroGen} & 72.04 & 79.32 & 18.11 & \cellcolor[rgb]{0.929,0.929,0.929}{\textbf{22.12}} & 12.22 & 76.42 & 83.01  & 19.11 & \cellcolor[rgb]{0.929,0.929,0.929}{\textbf{31.75}} & 13.04 & 9.8 $\times$ \\
\quad -\texttt{TDW} & 71.87 & 78.90  & 18.08 & \cellcolor[rgb]{0.929,0.929,0.929}{21.87} & 11.75 & 76.29 & 82.52  & 19.14 & \cellcolor[rgb]{0.929,0.929,0.929}{29.88} & 12.53 & 10.4 $\times$  \\
\quad -\texttt{VDW} & 75.44 & 82.06 & 17.56 & \cellcolor[rgb]{0.929,0.929,0.929}{20.50} & 11.62 & 77.80 & 83.84  & 18.20 & \cellcolor[rgb]{0.929,0.929,0.929}{29.63} & 12.62 & 11.7 $\times$ \\
\quad -\texttt{DW}        & 81.70 & 86.38 & 17.22 & \cellcolor[rgb]{0.929,0.929,0.929}{19.00} & 12.62 & 77.73 & 83.60 & 18.19 & \cellcolor[rgb]{0.929,0.929,0.929}{29.13} & 12.13 & 12.4 $\times$ \\
\quad -\texttt{T}$^*$     & 95.27  & 95.80 & \textbf{21.19} & \cellcolor[rgb]{0.929,0.929,0.929}{-}     & 10.84 & 95.27  & 95.80 & \textbf{21.19} & \cellcolor[rgb]{0.929,0.929,0.929}{-}     & 10.84    & 17.9 $\times$    \\
\textsc{ZeroGen} w/ obj & 81.56 & 87.93 & 19.42 & \cellcolor[rgb]{0.929,0.929,0.929}{16.37} & 12.93 & 82.23 & 87.93  & 19.66 & \cellcolor[rgb]{0.929,0.929,0.929}{29.76} & 13.30 & 9.8 $\times$ \\
\bottomrule[1pt]
\end{tabular}
\caption{Results of controllable news generation on \texttt{VisNews}. With $\textbf{C}_T$ controlling the sentiment, we regard the textual and the visual control as vertical elements in this task. Methods with * cannot be controlled w.r.t. sentiment.}
\label{tab:ctg_news}
\end{table*}

\begin{table}[!t]
\centering
\setlength\tabcolsep{1pt}
\small
\begin{tabular}{l|ccc|ccc} 
\toprule[1pt]
\multirow{2}{*}{Model} & \multicolumn{3}{c|}{Positive} & \multicolumn{3}{c}{Negative}  \\ 
                       & Flue.$\uparrow$ & Relv.$\uparrow$ & Sent.$\uparrow$/$\downarrow$   & Flue.$\uparrow$ & Relv.$\uparrow$ & Sent.$\uparrow$/$\downarrow$         \\ 
\midrule
\textsc{MAGIC}         &3.37   &2.77   &28.7/22.0   &\textbf{3.85}           &\textbf{3.13}  &46.0/14.7       \\
\quad +\textsc{PPLM}   &2.24   &2.85   &34.0/\textbf{7.3}   &3.12           &3.11  &52.0/10.7       \\
\midrule
\textsc{ZeroGen}       &\textbf{3.38}   &\textbf{2.94}   &\textbf{80.0}/10.7   &3.80           &2.85  &\textbf{84.7}/\textbf{6.0}        \\
\bottomrule[1pt]
\end{tabular}
\caption{Human evaluation results. Sent. scores are percentages of news that obeys/disobeys given sentiment.}
\label{tab:human_eval}
\end{table}

\subsection{Stylized Captioning}
To explore the sufficiency of our model to adapt to different styles, such as ``romantic'' or ``humorous''. We follow \citet{nukrai2022text} to conduct stylized-text fine-tuning in base LM for stylized captioning.

\paragraph{Dataset and Metrics.} In this task we still take textual objects from images as $\textbf{C}_T$ and we follow the exact experimental setting in previous works \cite{zhao2020memcap,nukrai2022text} on \texttt{FlickrStyle10k} dataset \cite{gan2017stylenet}. As for metrics, we take the same ones in Sec.~\ref{sec:image_caption}. Refer to Appendix~\ref{app:details_scd} for more detailed settings.
\paragraph{Main Results.} Table~\ref{tab:style_caption} shows quantitative results of the task, (1) \textsc{ZeroGen} outperforms most baselines on two stylized data, including weakly supervised \textsc{CapDec} on \texttt{Romantic} and \textsc{MemCap} with task-oriented training. (2) While it under-performs \textsc{CapDec} on some metrics over \texttt{Humorous} data, our method produces more fluent and plausible captions with consistently higher B@3 scores. (3) From two stylized sets, textual guidance takes a large credit to boost the model performance (comparing model -\texttt{T}), verifying the effectiveness of the proposed multimodal guidance in \textsc{ZeroGen}.

\subsection{Controllable News Generation} \label{sec:cng}
Textual guidance can not only serve as a complement to visual knowledge but can also be a lateral extension. In this task, we assign textual control for news sentiment guidance and visual control for image-relevant news generation.

\paragraph{Dataset and Metrics.} We conduct experiments on \texttt{VisNews} \cite{liu2020visual}. We fine-tune the base LM (i.e., SimCTG) on news data with the news title as an input prompt. We follow \citet{dathathri2019plug} to obtain word lists for two types of sentiment guidance respectively. We take four aspects for evaluation, diversity: Distinct-2 (D-2) and Distinct-4 (D-4) \cite{li2016diversity}. Image-text relevance: CLIP score (C-S) is the image-text similarity calculated by a CLIP. Control degree: $\Delta \text{Acc}$ (\%) evaluates the accuracy gain between generated sentences and human written news.\footnote{\textit{Human} written news in the test set consists of $62.88\%$ positive content and $37.12\%$ negative content.} Fluency: perplexity (PPL) measures the model output confidence.\looseness=-1

For human evaluation, we take Fluency (Flue.) for content fluency evaluation, Relevance (Relv.) for image/title-news relevance evaluation, and Sentiment (Sent.) to measure sentiment control. We strictly obey a double-blind procedure, where three annotators know nothing about the models. We sample 100 instances across every model.\footnote{Details of automatic metrics and human evaluation settings are in Appendix~\ref{app:details_cng} and \ref{sec:app_human_eval} respectively.}

\paragraph{Main Results.} From results in Table~\ref{tab:ctg_news}, we can draw the following conclusions: (1) \textsc{ZeroGen} has the highest classification accuracy gain and competitive CLIP scores among all presented statistics, proving that the proposed method can successfully produce controllable outputs under both modal supervisions. But our model generally degenerates the diversity, which we consider a trade-off. (2) Introducing dynamic weighting enhances the overall model performance. While \texttt{VDW} augments connections between the given image and generated news content with higher CLIP scores (C-S), \texttt{TDW} is able to make the output more recognizable w.r.t. the sentiment control without sacrificing content diversity. These findings validate the vastness and efficacy of the dynamic weighting mechanism even when their functional domains (i.e., $\textbf{C}_T, \textbf{C}_V$) are not complementary. (3) External controls jeopardize our model’s output confidence with slightly higher PPL than \textsc{MAGIC}, yet \textsc{ZeroGen} still largely outperforms \textsc{MAGIC+PPLM}, the only controllable counterpart on PPL. Also, \textsc{ZeroGen} without parts of the dynamic weighting (e.g., -\texttt{VDW}) can advantageously outgain \textsc{MAGIC+PPLM} on both controllability and diversity metrics. (4) \textsc{ZeroGen} requires no task-oriented training, thus registering its superiority in decoding efficiency over \textsc{MAGIC+PPLM} by nearly 10 times faster. 

We also present human evaluation results in Table~\ref{tab:human_eval},\footnote{The average Fleiss's Kappa \cite{fleiss1973equivalence} is 0.28, indicating three annotators reached a fair agreement.} which can further verify our findings above. 

\begin{table}[!t]
\centering
\small
\renewcommand\arraystretch{1}
\begin{tabular}{c|l|ccc} 
\toprule[1pt]
\multicolumn{2}{c|}{$\hat{\alpha}$}    & 5.0   & 8.0   & 10.0   \\ 
\midrule
\multirow{5}{*}{Pos.} & D-2 $\uparrow$ & \textbf{91.60}  & 72.04  & 52.37   \\
                     & D-4 $\uparrow$ & \textbf{95.00} & 79.32 & 60.00  \\
                     & C-S $\uparrow$ & \textbf{19.96} & 18.11 & 17.19  \\
                     & $\Delta \text{Acc} \uparrow$ & 10.75 & 22.12 & \textbf{26.62}  \\
                     & PPL $\downarrow$ & 11.76 & 12.22 & \textbf{10.28}  \\ 
\midrule
\multirow{5}{*}{Neg.} & D-2 $\uparrow$ & \textbf{92.95} & 76.42  & 52.09   \\
                     & D-4 $\uparrow$ & \textbf{95.87} & 83.01 & 60.07  \\
                     & C-S $\uparrow$ & \textbf{21.59} & 19.11 & 18.81  \\
                     & $\Delta \text{Acc} \uparrow$ & 11.26 & 31.75 & \textbf{51.26}  \\
                     & PPL $\downarrow$ & 11.71 & 13.04 & \textbf{11.52}  \\
\bottomrule[1pt]
\end{tabular}
\caption{Effect of different $\hat{\alpha}$ on news generation task.}
\label{tab:ctg_beta_upper}
\end{table}
\paragraph{Effect of $\boldsymbol{\alpha}$ Upper Bound.} $\textbf{C}_T$ is the only source for sentiment manipulation in this task, the upper bound of which can decide how distinguishable an output sentence is w.r.t. sentiment. In Table~\ref{tab:ctg_beta_upper}, with the increase of $\hat{\alpha}$, both accuracy and fluency will gain significant benefits. However, the diversity of texts and image relevance indicators will fall precipitously. This phenomenon is explained as more guiding information from sentiment may make the model inclined to only express the desired sentiment, trading for wider imagination associated with image or diversity. At the user end, we can twitch $\hat{\alpha}$ according to tasks to fit in different situations.

\paragraph{$\boldsymbol{\textbf{C}_T}$ Plays Two Roles.} Now we have examined the function of $\textbf{C}_T$ as a complement (captioning) or additional control element (news generation). We wonder \textit{can $\textbf{C}_T$ play both roles well at the same time?} We conduct experiments using our \textsc{ZeroGen} with both objects from the image and sentiment words as textual guidance. Our method with objects as a part of $\textbf{C}_T$ is marked with ``w/ obj''. Though \textsc{ZeroGen} w/ obj reaches a higher CLIP score, the accuracy and PPL generally decline comparing methods without textual objects guidance, which can also be accomplished by switching $\hat{\alpha}$ as mentioned earlier. That is to say, $\textbf{C}_T$ may be confused to play both roles as a complement and a lateral extension of images at the same time.

\begin{figure}[!t]
\centering
\includegraphics[width=1\linewidth]{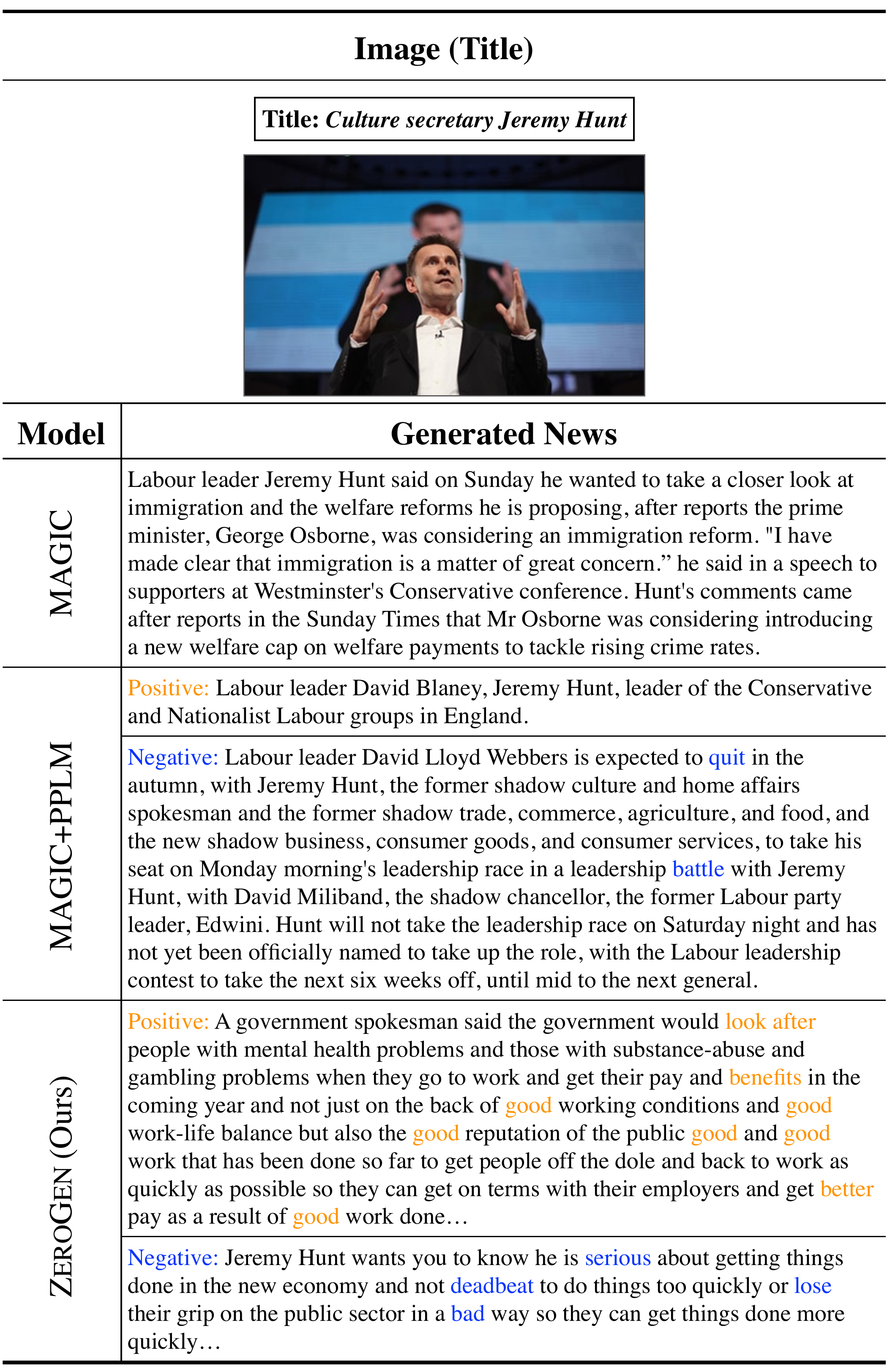}
\caption{An example of news from varied models. We highlight \textcolor{orange}{Positive} and \textcolor{blue}{Negative} words respectively.}
\vspace{-0.1in}
\label{fig:case_study}
\end{figure}

\paragraph{Case Analysis.} We exhibit an example of generated controllable news in Figure~\ref{fig:case_study}. As the image shows \textit{Culture secretary Jeremy Hunt} is giving a talk. All methods are able to produce image/title-relevant sentences, but \textsc{MAGIC+PPLM} generates some false evidence such as recognizing Jeremy Hunt as the ``leader of Conservative and Nationalist Labour groups''. Besides, our \textsc{ZeroGen} can produce more diverse and controllable words like ``good'', ``benefits'' for positive and ``deadbeat'', ``lose'' for negative, while \textsc{MAGIC+PPLM} is not competent to fulfill the controllable aspect. More cases are exhibited in Appendix~\ref{app:cases}.

\section{Conclusion}
In this paper, we present \textsc{ZeroGen}, a paradigm of zero-shot controllable text generation with multimodal signals. We explicitly separate visual control and textual control into sentence-level and token-level guidance. And we use two \textit{Oracle}s to unify the control signals to LM output probability space. A dynamic weighting mechanism is applied to adapt to all multimodal controls, which further boosts the model generation ability. Three tasks from captioning to controllable news generation justify the effectiveness of \textsc{ZeroGen} and help us explore the relationship between distinct signals. By providing multimodal knowledge, we demonstrate LMs without task-specified training can substantially achieve astonishing performance in multimodal tasks across different setups and domains. \looseness=-1

\section{Limitations}
Although \textsc{ZeroGen} successfully achieves zero-shot controllable text generation, our technique is still subject to a few limitations to be addressed in follow-up work. (1) There is still a large gap between weakly and fully supervised methods. We believe the rich semantic information contained in these large-scale pre-trained \textit{Oracle}s we employed can further narrow it. (2) The diversity in our controllable news generation task is insufficient. Since this is a widespread problem in the field of zero-shot research, we plan to alleviate the issue by incorporating more diverse language decoding schemes \cite{xu2022learning} or partial training parameters in the model such as adapters \cite{houlsby2019parameter}. (3) The existence of spurious correlations \cite{tu2020empirical,chai2022fast} in bad cases (as shown in Appendix~\ref{app:cases}) is nonnegligible, one of our future work directions is to handle it by introducing causal inference \cite{pearl2009causal}.

\section{Ethics Statement}
We are well aware that text generation technologies may be abused to create deceptive, harmful, or objectionable content. For our \textsc{ZeroGen}, we can conduct experiments on detoxification datasets \cite{gehman2020realtoxicityprompts} to make it a useful tool for combating hate speech and eliminating harmful information in PLMs. As we are considering components to make our method more robust and effective in multimodal controllable tasks, we believe it is meaningful and beneficial to progress research on controllable text generation.

\bibliography{anthology,custom}
\bibliographystyle{acl_natbib}

\newpage
\appendix

\section{Implementation Details} \label{app:details}
\subsection{General Model Details}
For the base LM, we use SimCTG \cite{su2022contrastive}, which is extended from a pre-trained GPT-2 \cite{radford2019language} model. SimCTG essentially consists of contrastive training and contrastive searching. As for training period, it introduces a $\mathcal{L}_{\mathrm{CL}}$ term to learn discriminative and isotropic token representations:
\begin{equation}
\begin{aligned}
\mathcal{L}_{\mathrm{CL}}&=\frac{1}{V \times(V-1)} \sum_{i=1}^{V} \sum_{j=1, j \neq i}^{V} \max \{0, \\& \rho-s\left(h_{x_i}, h_{x_i}\right)+s\left(h_{x_i}, h_{x_j}\right)\},
\end{aligned}
\end{equation}
where $V$ is the vocabulary size, function $s(\cdot, \cdot)$ is the similarity function, $h_{x_i}$ is the LM hidden state of token $x_i$ and $\rho$ is a pre-defined margin. The final training objective of a LM turns into:

\begin{equation}
    \mathcal{L}_{\text{SimCTG}} = \mathcal{L}_{\mathrm{MLE}} + \mathcal{L}_{\mathrm{CL}},
\end{equation}
with $\mathcal{L}_{\mathrm{MLE}}$ to be the vanilla MLE objective of LM.

As for contrastive decoding, at decoding time $t$, the token to be generated is formalized as:
\begin{equation}
    \begin{aligned} \nonumber
    &S_{\text{SimCTG}}(x_{<t}) = (1-\eta) \times \underbrace{p_\theta\left(w \mid x_{<t}\right)}_{\text{model confidence}}- \\& \eta \times \underbrace{\left(\max \left\{s\left(h_w, h_{x_j}\right): 1 \leq j \leq t-1\right\}\right)}_{\text{degeneration penalty}}\\& 
    x_t =\underset{w \in W_t^{(k)}}{\arg \max } S_{\text{SimCTG}}(x_{<t})
    \end{aligned}
\end{equation}
where $\eta$ is a parameter to balance generation diversity and consistency. Then for our \textsc{ZeroGen}, we have the following decoding objective based on the shifted LM output $p'_{\text{LM}_t}$:
\begin{equation}
    \begin{aligned} \nonumber
    x_t &=\underset{w \in W_t^{(k)}}{\arg \max } \{S_{\text{SimCTG}}(x_{<t}) \\&+ \beta_t \times S_{t}(w, \textbf{C}_V\mid x_{<t})\},
    \end{aligned}
\end{equation}
here $W_t^{(k)}$ is grouped from $p'_{\text{LM}_t}$ and $S_t$ is the \textsc{MAGIC} score we introduced in Sec.~\ref{sec:visual_control_method}.


When we apply \textsc{ZeroGen}, there are several parameters should be decided in advance: $k$ in $W_t^{(k)}$, $\eta$ for contrastive decoding, $\beta, \alpha, \hat{\beta}, \hat{\alpha}$ for dynamic weighting mechanism. We present detailed parameters in Table~\ref{tab:params} to aid reproducibility. We also present the workflow of our system in Algorithm~\ref{alg:main}.

We implement all the experiments on the same machine with one NVIDIA GeForce RTX 3090 GPU with 24G memory and one Intel 3.70GHz i9-10900K CPU. We will release the code of all methods (including baselines) and datasets processing once the paper is accepted.

\begin{algorithm}[!t]
\caption{\textsc{ZeroGen}} 
\hspace*{0.02in} {\bf Input:}
{
Visual control: $\textbf{C}_V$, textual control: $\textbf{C}_T$
}\\
\hspace*{0.02in} {\bf Output:} 
Generated content $\textbf{X} = [x_1, x_2, ...]$
\begin{algorithmic}[1]
\State initialize $\textbf{V}$; \texttt{//LM vocabulary}
\State initialize $\phi_M, \phi_T$; \texttt{//oracles}
\State initialize $\hat{\beta}, \hat{\alpha}, k, \lambda$; \texttt{//hyper params}
\State compute $p(\textbf{V}, \textbf{C}_T)$ using $\phi_T$;
\State $x_0 \longleftarrow \texttt{[BOS]}, \textbf{X} \longleftarrow \text{[}x_0\text{]}, t \longleftarrow 0$;
\While{$x_t \not = \texttt{[EOS]}$}
    \State $t\leftarrow t+1$;
    \State compute $p_{\text{LM}_t}$ using base LM and $x_{<t}$;
    \State compute $p_t(\textbf{C}_T)$ using $p(\textbf{V}, \textbf{C}_T)$;
    \State $D_T \leftarrow \sum_{n} p_{\text{LM}_t}(C_{T_n}\mid x_{<t}) / N$;
    \State $\alpha_t \leftarrow \min (D_T / \lambda, \hat{\alpha})$;
    \State $p'_{\text{LM}_t} \leftarrow p_{\text{LM}_t} + \alpha_t \times p_t(\textbf{C}_T)$;
    \State $D_V \leftarrow \sum_{w\in W_t^{(k)}}p_{\phi_M}(w_t, \textbf{C}_V) / k$;
    \State $\beta_t \leftarrow \min ( D_V / \lambda, \hat{\beta})$;
    \State compute $S_{t}(w, \textbf{C}_V\mid x_{<t})$ using $\phi_M$;
    \State $x_t \leftarrow \arg\max_{w}S_{t}(w, \textbf{C}_V\mid x_{<t})$;
    \State add $x_t$ to content $\textbf{X}$;
\EndWhile
\State \textbf{return} generated content \textbf{X};
\end{algorithmic}
\label{alg:main}
\end{algorithm}

\begin{table}
\centering
\setlength\tabcolsep{3pt}
\small
\begin{tabular}{l|ccccc} 
\toprule[1pt]
    Dataset & Train  & Val  & Test & \# Voc    & \# Len  \\ 
\midrule
\texttt{F10k Humor}    & 6,000   & 1,000 & 1,000 & 7,186  & 14.07    \\
\texttt{F10k Romantic} & 6,000   & 1,000 & 1,000 & 6,434  & 14.55    \\
\texttt{VisNews}      & 13,098 & 200  & 800  & 23,274 & 136.20   \\
\bottomrule[1pt]
\end{tabular}
\caption{Detailed statistics of data employed in our tasks. \# Voc and \# Len represent the vocabulary size and average sentence length of current dataset. \texttt{F10k} represents the \texttt{Flickr10k} dataset.}
\label{tab:data_details}
\end{table}

\begin{table*}
\centering
\small
\begin{tabular}{l|ccccc} 
\toprule[1pt]
\# Params / Data & \texttt{MS-COCO}   & \texttt{Flickr30k}  & \texttt{F10k Romantic} & \texttt{F10k Humor} & \texttt{VisNews}  \\ 
\midrule
$k$ (\texttt{int})         & 45                  & 25                  & 45            & 45         & 5              \\
$\eta$ (\texttt{float})       & 0.10                 & 0.10                 & 0.10           & 0.10        & 0.65           \\
$\alpha$ (\texttt{float})         & 1.0                 & 2.0                 & 1.0           & 1.0        & 8.0             \\
$\beta$ (\texttt{float})        & 1.0                 & 1.0                 & 1.0           & 1.0        & 1.0             \\
$\hat{\alpha}$ (\texttt{float})      & 2.5                 & 2.0                 & 3.0           & 2.5        & 8.0            \\
$\hat{\beta}$ (\texttt{float})        & 1.0                 & 0.5                 & 0.5           & 0.5        & 0.5             \\
$N$ (\texttt{int})        & 1$\sim$5 & 1$\sim$5  & 1             & 1          & 40             \\
\bottomrule[1pt]
\end{tabular}
\caption{Detailed parameters of \textsc{ZeroGen} for different tasks.}
\label{tab:params}
\end{table*}

\subsection{Image Captioning Details}
For both \texttt{MS-COCO} and \texttt{Flickr30k} datasets, we take the Karpathy split \cite{karpathy2015deep} and use the train, valid sets for base LM trainin and test set for the task evaluation. In detail, \texttt{MS-COCO} dataset is under a Creative Commons Attribution 4.0 License, and is publicly available at \url{https://cocodataset.org}, while \texttt{Flickr30k} is under a Creative Commons Public Domain Dedication License, and is publicly available at \url{https://www.kaggle.com/datasets/hsankesara/flickr-image-dataset}. For the base LM, we load the publicly available pre-trained language model weights\footnote{\url{https://huggingface.co/cambridgeltl/magic_flickr30k} and \url{https://huggingface.co/cambridgeltl/magic_mscoco}} and set the maximum decoding length to be 16 for this task.

\subsection{Stylized Captioning Details} \label{app:details_scd}
For stylized captioning and controllable news generation task, the \texttt{Flickr10k Stylized} dataset is introduced by \citet{gan2017stylenet} and under an unknown license, it can be publicly downloaded from \url{https://zhegan27.github.io/Papers/FlickrStyle_v0.9.zip}. On model side, we follow \citet{zhao2020memcap} to randomly sample 6,000 instances from the original corpus as training data and 1,000 as test data. Their detailed statistics are shown in Table~\ref{tab:data_details}. Following \citet{nukrai2022text}, we fine-tune two base LMs on two training sets respectively to achieve stylized outputs. As for the base LM fine-tuning, we use SimCTG with 0.5 to be the margin $\rho$ and 1e-5 as the learning rate to train the model until it has no further loss decrease on the valid set. We set the maximum sentence length to 128 for base LM training and 25 for decoding.

\subsection{Controllable News Generation Details} \label{app:details_cng}
For dataset and metrics, we conduct experiments based on \texttt{VisNews} \cite{liu2020visual}. This dataset is under an unknown license, and can be acquired by asking the author directly.\footnote{\url{https://github.com/FuxiaoLiu/VisualNews-Repository}} Specifically, for image-text data, we sampled 13000, 200, and 800 image-news pairs from the original \texttt{VisNews} dataset as train, valid, and test set. We use train and valid set for base LM (i.e., SimCTG) fine-tuning with the news title as an input prompt. The test set is employed for the final evaluation. Details are shown in Table~\ref{tab:data_details}. The max training news length is 200, and the max decoding length is set to 130.

\begin{figure}[!t]
\centering
\includegraphics[width=1\linewidth]{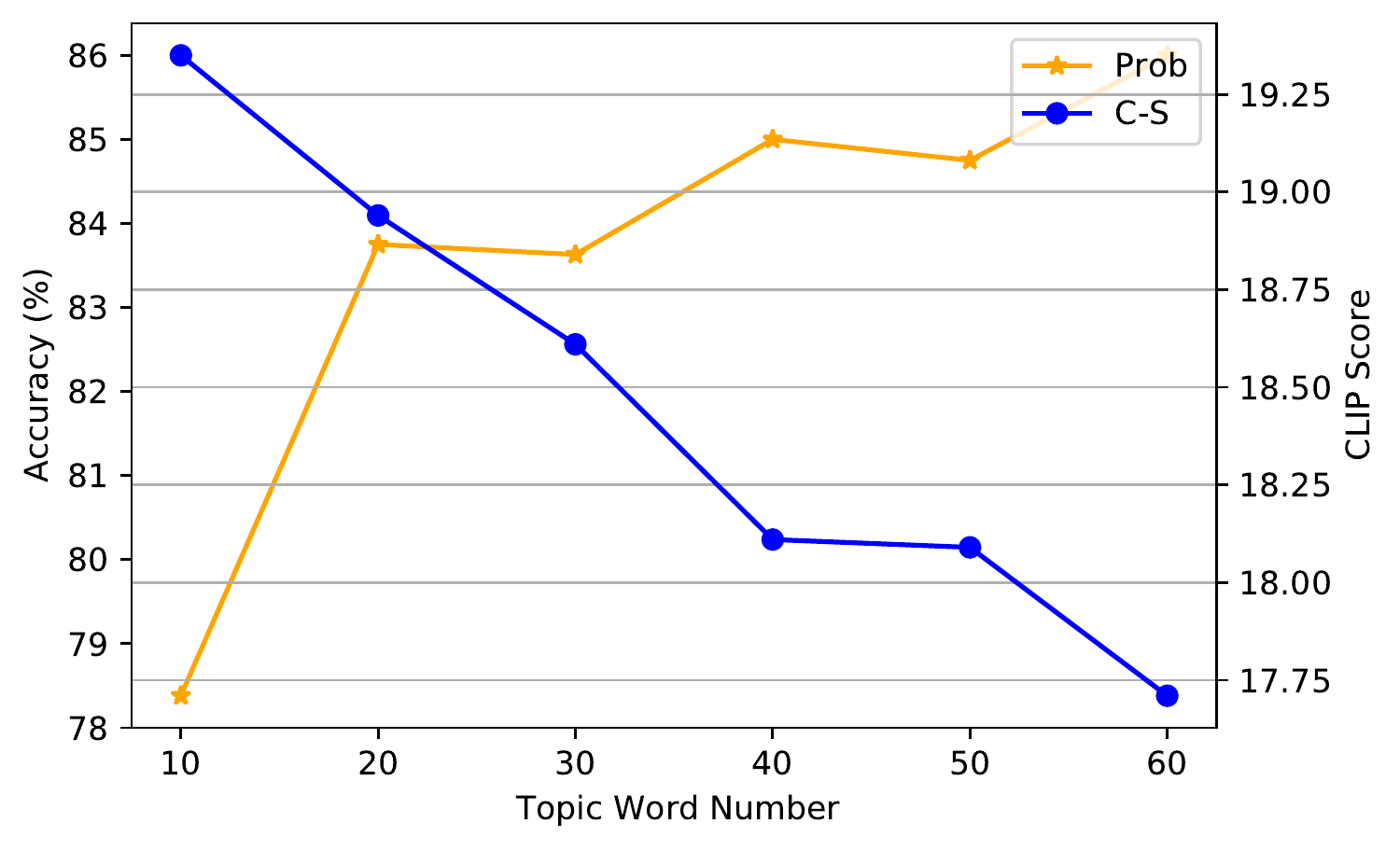}
\caption{Classification accuracy (Prob) and CLIP-score (C-S) with varied topic word number $N$.}
\vspace{-0.1in}
\label{fig:twn_abla}
\end{figure}

For word bags of two sentiments, we follow \citet{dathathri2019plug} to obtain word lists of ``happiness'' and ``negative words'' for positive and negative sentiment guidance respectively.\footnote{Word lists are downloaded from \url{www.enchantedlearning.com/wordlist}.} As for evaluation, diversity, we use the CLIP model that is different from the one guides multimodal generation in \textsc{ZeroGen} to compute the CLIP score (C-S). For $\Delta \text{Acc}$ (\%), we take the pre-trained SiEBRT model \cite{hartmann2022more} that has SOTA performance on SST-2 dataset \cite{socher2013recursive} as the classifier.

From Table~\ref{tab:params}, we can notice that we select the topic number $N$ to be 40 in this task, meaning at every decoding time, we input 40 sentiment words as textual control $\textbf{C}_T$. We conduct ablation experiments w.r.t. topic number $N$ and the classification accuracy as well as CLIP score in Figure~\ref{fig:twn_abla}. From the picture, we can figure out that, as the increase of $N$, the control degree gradually gets higher with better accuracy while the image-news relevance declines with lower C-S. This is because more words from one sentiment as $\textbf{C}_T$ make the model more focused on sentiment control but sacrifice some image-related text generative ability. 

\begin{table*}[!t]
\setlength\tabcolsep{3pt}
\centering
\small
\begin{tabular}{l|cccccc|cccccc} 
\toprule[1pt]
\multirow{2}{*}{Model}      & \multicolumn{6}{c|}{\texttt{MS-COCO}}                                                                                   & \multicolumn{6}{c}{\texttt{Flickr30k}} \\ 
\cmidrule{2-13}
                            & B@1 $\uparrow$ & B@4 $\uparrow$ & M $\uparrow$  & R-L $\uparrow$ & CIDEr $\uparrow$ & SPICE $\uparrow$         & B@1 $\uparrow$ & B@4 $\uparrow$ & M $\uparrow$  & R-L $\uparrow$ & CIDEr $\uparrow$ & SPICE $\uparrow$                          \\ 
\midrule
\textsc{ZeroGen} ($N=1$)              & 59.4  & 15.5  & \textbf{18.7} & 42.3  & 55.4    & \textbf{12.1}         & 54.9  & 13.1  & 15.2 & 37.4  & 26.4    & \textbf{8.3}           \\
\textsc{ZeroGen} ($N=2$)              & 60.1  & 15.6  & 18.5 & 42.3  & 55.9    & 11.9         & \textbf{55.3}  & \textbf{13.3}  & \textbf{15.4} & \textbf{37.7}  & \textbf{27.5}    & \textbf{8.3}           \\
\textsc{ZeroGen} ($N=3$)              & 60.4  & 15.6  & 18.6 & 42.3  & 56.5    & \textbf{12.1}         & 55.0  & 13.1  & 15.2 & 37.4  & 27.1    & 8.2           \\
\textsc{ZeroGen} ($N=4$)              & 60.5  & 15.7  & \textbf{18.7} & \textbf{42.4}  & 57.0    & \textbf{12.1}         & 54.5  & 13.1  & 15.2 & 37.5  & 27.2    & \textbf{8.3}           \\
\textsc{ZeroGen} ($N=5$)              & \textbf{60.6}  & \textbf{15.8}  & \textbf{18.7} & \textbf{42.4}  & \textbf{57.1}    & \textbf{12.1}         & 55.0  & 13.0  & 15.2 & 37.5  & 27.3    & 8.2           \\

\bottomrule[1pt]
\end{tabular}
\caption{Caption results of \textsc{ZeroGen} with only varied number of objects as textual control for each generation.}
\label{tab:full_N_captioning}
\end{table*}
\begin{table*}[!t]
\setlength\tabcolsep{3pt}
\centering
\small
\begin{tabular}{l|cccccc|cccccc} 
\toprule[1pt]
\multirow{2}{*}{Model}      & \multicolumn{6}{c|}{\texttt{MS-COCO}}                                                                                   & \multicolumn{6}{c}{\texttt{Flickr30k}} \\ 
\cmidrule{2-13}
                            & B@1 $\uparrow$ & B@4 $\uparrow$ & M $\uparrow$  & R-L $\uparrow$ & CIDEr $\uparrow$ & SPICE $\uparrow$         & B@1 $\uparrow$ & B@4 $\uparrow$ & M $\uparrow$  & R-L $\uparrow$ & CIDEr $\uparrow$ & SPICE $\uparrow$                          \\ 
\midrule
\textsc{ZeroGen} \textbf{SR}   & 59.6  & 15.3  & 18.4 & 42.1  & 55.5    & 11.8         & 54.8  & 13.1  & 15.0 & 37.2  & 26.7    & 8.1           \\
\textsc{ZeroGen} \textbf{MP}       & 59.9  & 15.2  & 18.3 & 42.0  & 55.2    & 11.8         & 54.8  & 13.2  & 15.1 & 37.5  & 26.7    & 8.1           \\
\textsc{ZeroGen} \textbf{WM}      & \textbf{60.6}  & \textbf{15.8}  & \textbf{18.7} & \textbf{42.4}  & \textbf{57.1}    & \textbf{12.1} & \textbf{55.3}  & \textbf{13.3}  & \textbf{15.4} & \textbf{37.7}  & \textbf{27.5}    & \textbf{8.3}           \\

\bottomrule[1pt]
\end{tabular}
\caption{Caption results of \textsc{ZeroGen} with only three different $p_t(\textbf{C}_T)$ selection methods. \textbf{WM, MP, SR} represent Word-wise Max, Mean Pooling and Step-wise Random in Sec.~\ref{sec:textual-control} respectively.}
\label{tab:ct_selection}
\end{table*}
\begin{table*}[!t]
\setlength\tabcolsep{4pt}
\centering
\small
\begin{tabular}{l|cccccc|cccccc} 
\toprule[1pt]
\multirow{2}{*}{Model} & \multicolumn{6}{c|}{\texttt{MS-COCO} $\implies$ \texttt{Flickr30k}}             & \multicolumn{6}{c}{\texttt{Flickr30k} $\implies$ \texttt{MS-COCO}}             \\ 
\cmidrule{2-13}
                       & B@1 $\uparrow$  & B@4 $\uparrow$ & M $\uparrow$   & R-L $\uparrow$  & CIDEr $\uparrow$ & SPICE $\uparrow$ & B@1 $\uparrow$  & B@4 $\uparrow$ & M $\uparrow$   & R-L $\uparrow$  & CIDEr $\uparrow$ & SPICE $\uparrow$ \\ 
\midrule
\textsc{ZeroCap}                 & 49.2 & 6.2 & 11.9  & 29.3 & 18.3   & -   & 46.3   & 6.0   & 13.7    & 30.1   & \textbf{27.3}     & -      \\
\textsc{MAGIC}                  & 46.4 & 6.2 & 12.2 & 31.3 & 17.5  & 5.9   & 41.4 & 5.2 & 12.5 & 30.7 & 18.3  & 5.7    \\
\midrule
\textsc{ZeroGen}          & \textbf{50.5} & \textbf{8.1} & \textbf{13.1} & \textbf{34.5} & 17.3  & 6.0   & \textbf{46.9} & \textbf{7.6} & \textbf{14.0} & \textbf{34.4} & {26.1}  & \textbf{6.8}    \\
\quad -\texttt{TDW}        & 50.1 & 8.0 & 12.7 & 34.0 & 17.0  & 5.8   & 46.2 & 7.1 & 13.5 & 33.7 & 23.9  & 6.2    \\
\quad -\texttt{T} & 49.3 & \textbf{8.1} & 12.5 & 33.7 & 16.7  & 5.8   & 45.1 & 6.7 & 13.1 & 33.3 & 22.4  & 5.9    \\
\quad -\texttt{VDW}        & 49.3 & 7.2 & 13.0 & 33.5 & \textbf{18.5}  & \textbf{6.2}   & 43.8 & 6.2 & 13.5 & 32.6 & 24.5  & 6.3    \\
\quad -\texttt{DW}               & 48.2 & 7.1 & 12.5 & 32.8 & 17.4  & 5.9   & 43.7 & 6.2 & 13.5 & 32.6 & 24.4  & 6.3    \\
\bottomrule[1pt]
\end{tabular}
\caption{Cross-domain results on two image-caption datasets \texttt{MS-COCO} and \texttt{Flickr30k}.}
\label{tab:cross_domain}
\end{table*}

\subsection{Baseline Model Details} \label{app:baselines}
For \textsc{IC-SME}, \textsc{S2S-GCC} and \textsc{CapDec} results, we directly take them from their respect paper. For \textsc{ZeroCap}, we take their official implementation from \url{https://github. com/YoadTew/zero-shot-image-to-text} and use its default parameter setting for captioning tasks. For \textsc{MAGIC}, we take the official code from \url{https://github.com/yxuansu/MAGIC} to reproduce the results. 

For \textsc{MAGIC+PPLM} implementation, we take two codebases into consideration, including the official \textsc{PPLM} code from \url{https://github.com/uber-research/PPLM} and a simpler version of \textsc{PPLM} from \url{https://github.com/hit-scma/CAT-PAW}. And we add \textsc{MAGIC} process at the decoding stage of \textsc{PPLM} as \textsc{MAGIC+PPLM}, we provide a minimum reproducible code of it in this anonymous repository: \url{https://anonymous.4open.science/r/Pplm_Magic-3E15}. For positive sentiment control, we set the iteration for each LM hidden state update to be 5 with step size to be 0.03, while 15 iterations for negative control. And we choose (15+5)/2=10 iterations for decoding time measure in Sec.~\ref{sec:cng}. We use the same SimCTG model in \textsc{ZeroGen} on \texttt{VisNews} for \textsc{MAGIC+PPLM} generation. We set the max decoding length to be 130, $k=5$ for \textsc{MAGIC} searching in \textsc{MAGIC+PPLM} like our method. Other hyper-parameters are the default values in the official code repositories.

\section{Additional Experimental Results}
\subsection{Number of Objects in $\boldsymbol{\textbf{C}_T}$ for Captioning} \label{app:BoOct}
We display the full results of varying the number of objects in $\textbf{C}_T$ (number $N$) for captioning task in Table~\ref{tab:full_N_captioning}. For images with less than $N$ objects extracted, we just use their all existing objects as the textual control $\textbf{C}_T$. We can observe that, for both datasets, more textual guidance can bring performance gain. Nevertheless, increasing objects in this process do not necessarily gain better metrics. For instance, on \texttt{Flickr30k}, the model with $N=2$ performs the best among other $N$ settings. This may be because too much textual guidance cause confusion for \textsc{ZeroGen} on relatively easy tasks (i.e., shorter captions, smaller vocabulary size).
\subsection{$\boldsymbol{p(\textbf{C}_{T})}$ Selection Method in Captioning}
We also present full results of $p(\textbf{C}_{T})$ selection methods in Table~\ref{tab:ct_selection}. On two datasets, the results are consistent and indicate that \textbf{WM} is the best way for calculating $p_t(\textbf{C}_T)$ at $t$-th step.
\subsection{Image Captioning in Cross Domain} \label{app:cross_domain_cap}
For cross-domain captioning evaluation, we follow the setting in \citet{su2022contrastive} and fine-tune the base LM on one dataset (e.g., \texttt{MS-COCO}), while evaluating the model's captioning capacity on another dataset (e.g., $\texttt{MS-COCO}\implies$ \texttt{Flickr30k}). Results in Table~\ref{tab:cross_domain} can verify findings from in-domain evaluations in Sec.~\ref{sec:image_caption}.


\begin{figure}[!t]
\centering
\includegraphics[width=1\linewidth]{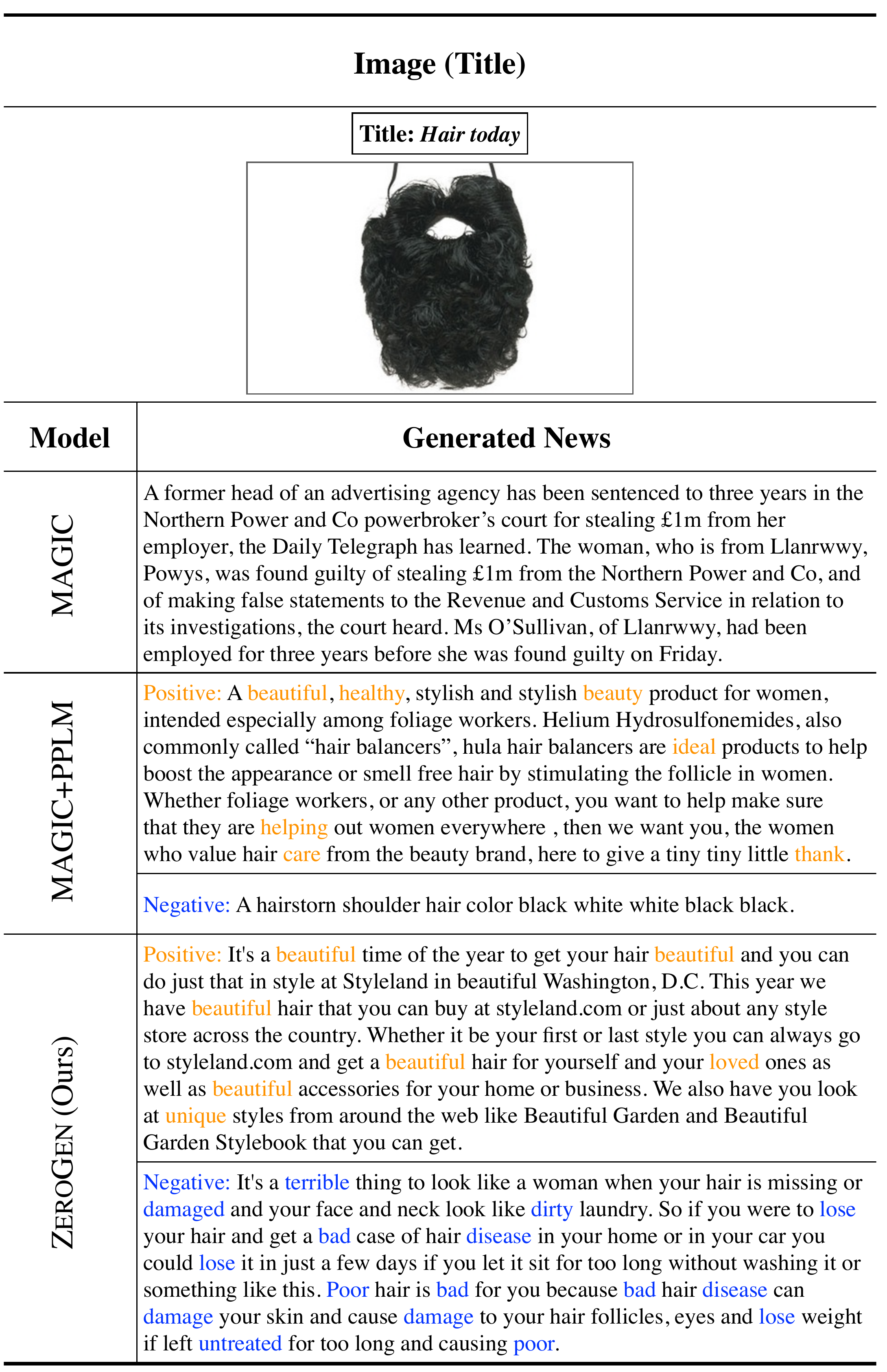}
\caption{Good case of news generated by our models comparing baselines.}
\vspace{-0.1in}
\label{fig:case_study2}
\end{figure}

\begin{figure}[!t]
\centering
\includegraphics[width=1\linewidth]{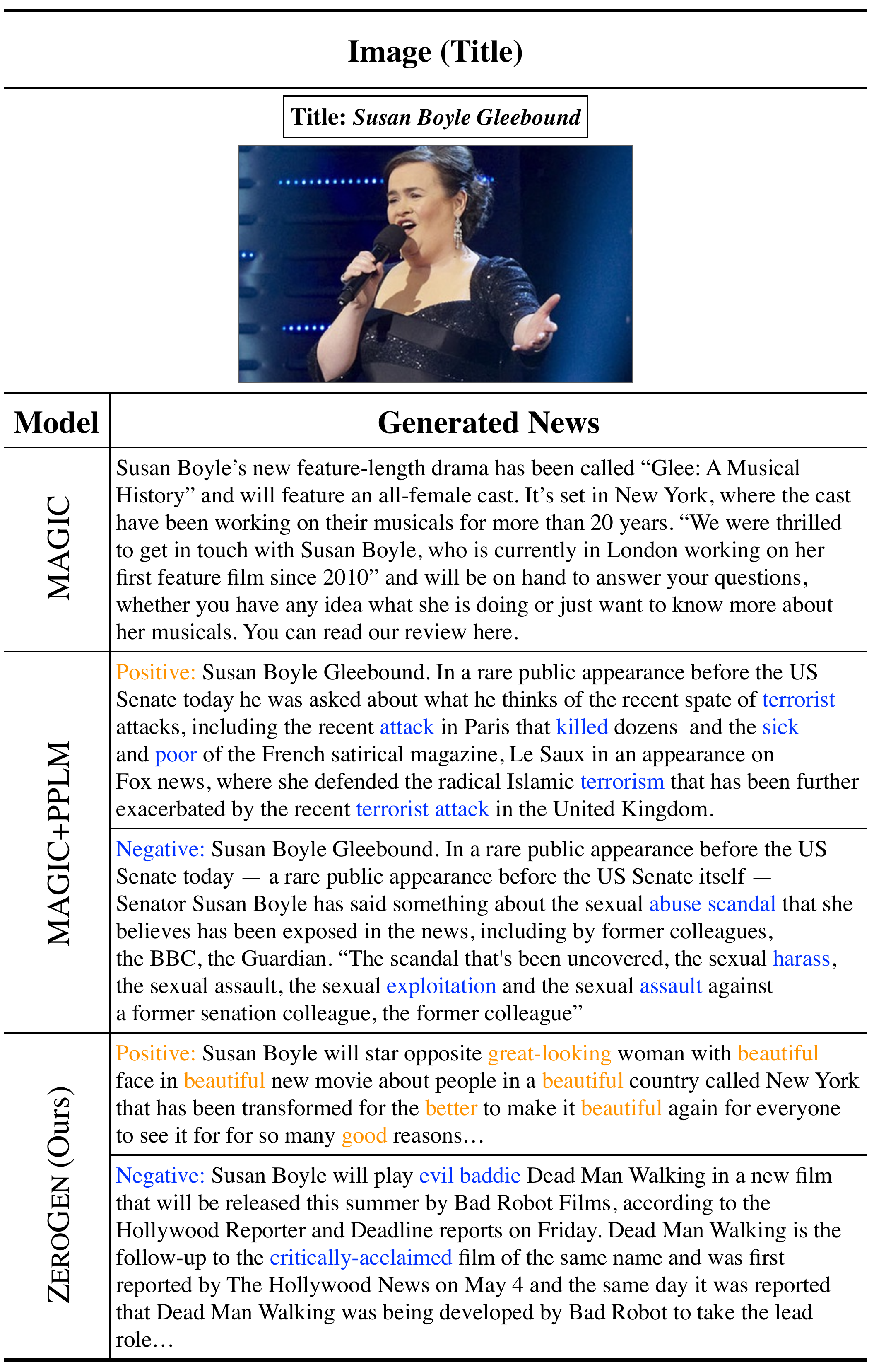}
\caption{Good case of news generated by our models comparing baselines.}
\vspace{-0.1in}
\label{fig:case_study3}
\end{figure}

\section{More Cases of \textsc{ZeroGen}} \label{app:cases}
In the presented cases, we highlight \textcolor{orange}{Positive} words and \textcolor{blue}{Negative} words respectively.
\paragraph{Good Cases.} As shown in Figure~\ref{fig:case_study2} and \ref{fig:case_study3}, the proposed method is capable to generate image-related and sentiment-controllable texts even with very few prompt words (Figure~\ref{fig:case_study2}). And our \textsc{ZeroGen} generally produce more diverse sentiment-specified words like ``beautiful'', ``unique'', ``great-looking'' and ``good'' for positive sentiment and ``terrible'', ``damaged'', ``dirty'' and ``evil'' for negative sentiment. The compared baselines are unable to generate controllable news. For instance, \textsc{MAGIC+PPLM} generates very short texts given negative sentiment for \textit{Hair today} image in Figure~\ref{fig:case_study2} and negative contents given positive sentiment in Figure~\ref{fig:case_study3}.

\begin{figure}[!t]
\centering
\includegraphics[width=1\linewidth]{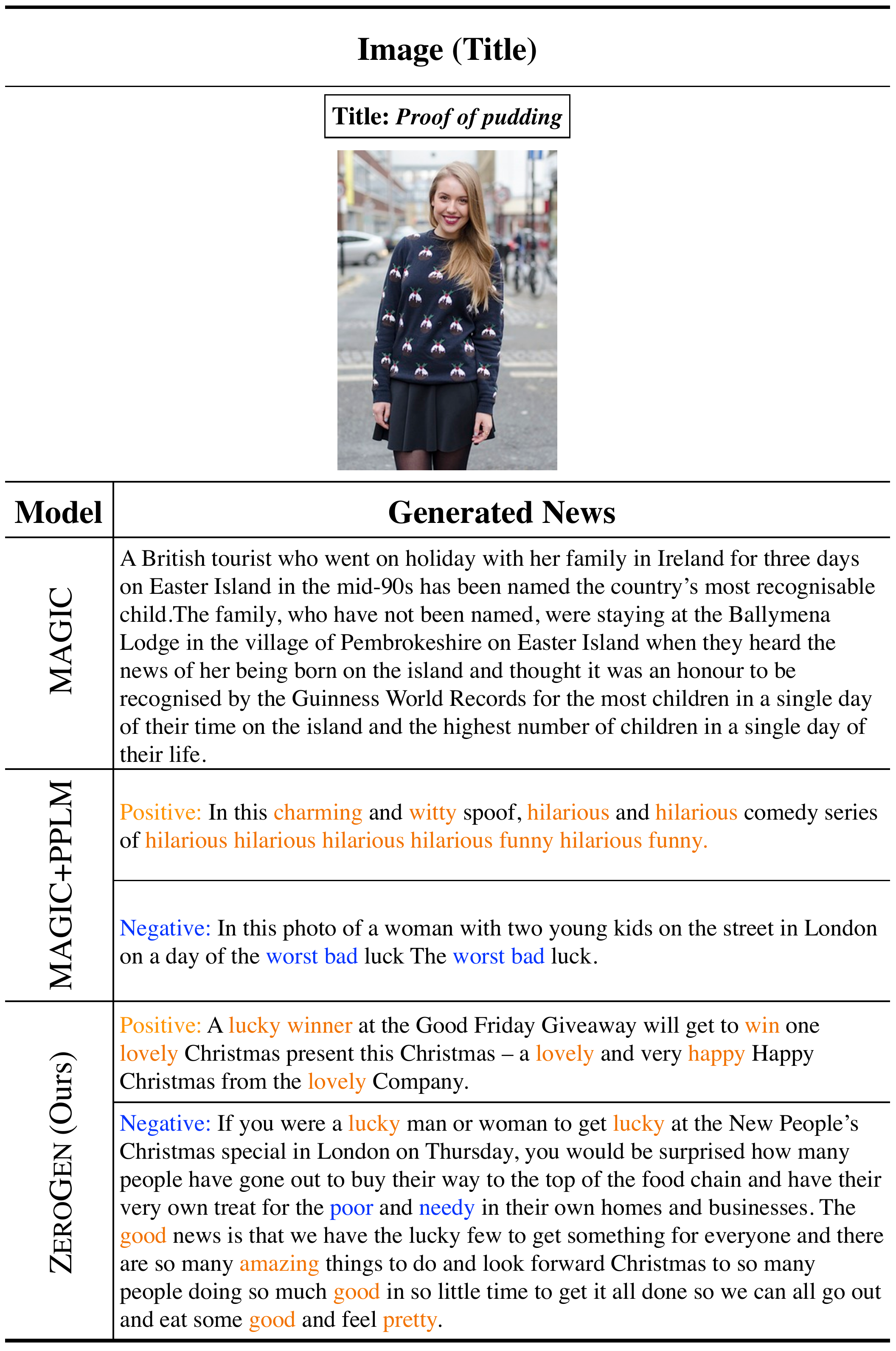}
\caption{Bad case of news generated by our models comparing baselines.}
\vspace{-0.1in}
\label{fig:bad_case_study1}
\end{figure}

\begin{figure}[!t]
\centering
\includegraphics[width=1\linewidth]{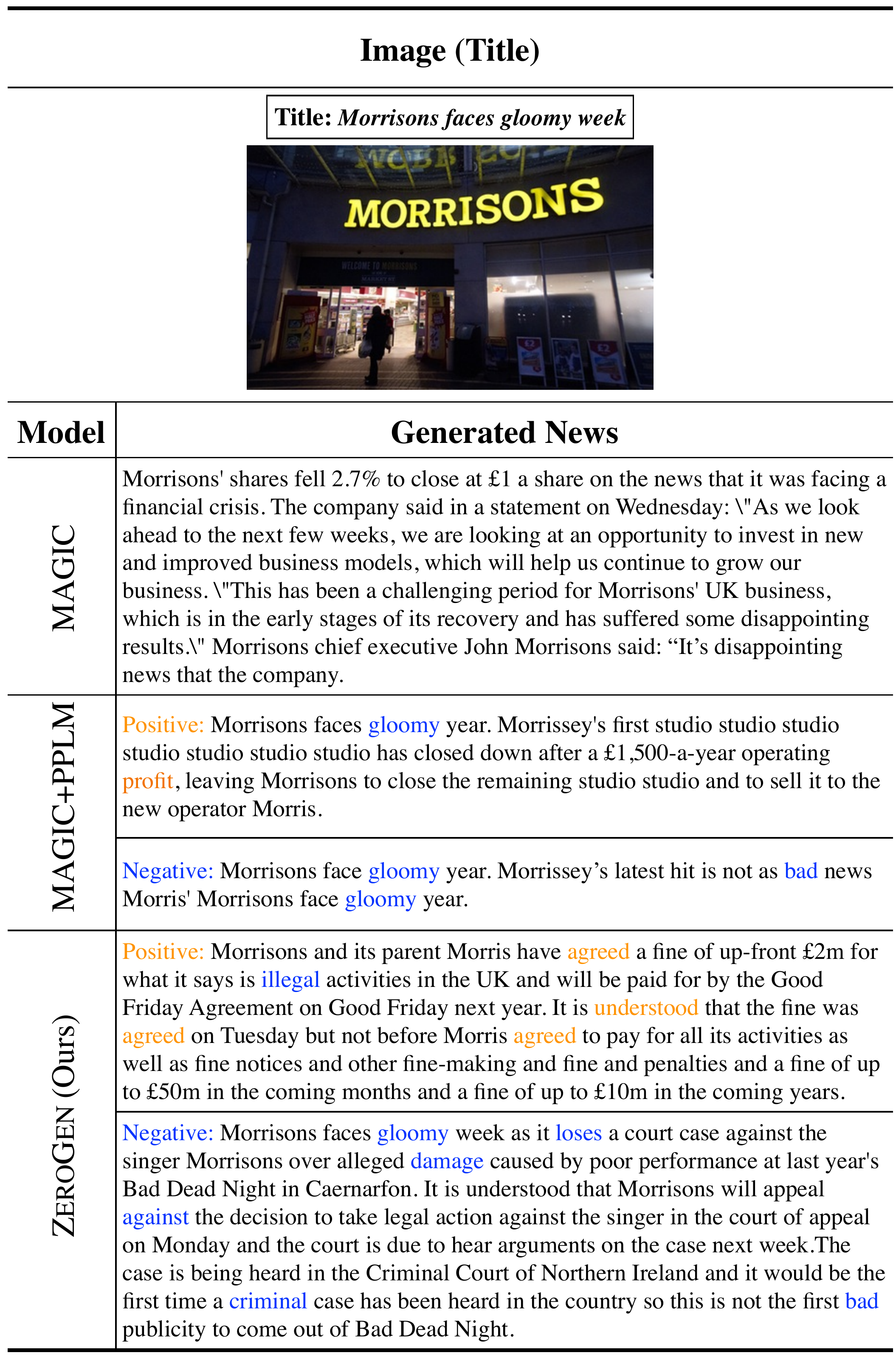}
\caption{Bad case of news generated by our models comparing baselines.}
\vspace{-0.1in}
\label{fig:bad_case_study2}
\end{figure}

\paragraph{Bad Cases.} We present bad cases of \textsc{ZeroGen} in Figure~\ref{fig:bad_case_study1} and \ref{fig:bad_case_study2}. For both cases, we can observe that there may exist generating biases caused by spurious correlation in dataset \cite{tu2020empirical,chai2022fast}. In Figure~\ref{fig:bad_case_study1}, the image describes a smiling woman with a flowered sweater, which means the visual control may be confounded with textual control when $\textbf{C}_T$ includes negative words (under the case where no task-oriented training is conducted). Our method struggles to generate negative-only content given the image and negative keywords. In Figure~\ref{fig:bad_case_study2}, the title \textit{Morrisons faces gloomy week} gives away its sentimental preference (i.e., negative) for the image. Similarly, for both \textsc{ZeroGen} and \textsc{MAGIC+PPLM}, they fail to generate news with only positive sentiment given positive textual control. We plan to take causal-related solutions such as self-training and causal intervention \cite{pearl2009causal,chai2022fast} towards this issue in the future.

\section{Human Evaluation} \label{sec:app_human_eval}
For annotators, we hire three graduate students from America or China with fluent English reading skills. Each annotator is assigned $100$ (instances)$\times 3$ (models)$\times 3$ (aspects) $=900$ rating tasks, resulting in $900$ (tasks)$\times 3$ (annotators) $=2,700$ human ratings in total. We use a three-scale scheme (i.e., integrate score 1, 2 to the poor class, 3 to the moderate class and 4, 5 to the good class) for Fluency and Relevance to compute the Fleiss's kappa \cite{fleiss1973equivalence}. The annotators have acknowledged the use of annotated data sets and are paid an average annotation salary. All annotators were aware of the potential risks or ethical concerns of machine-generated texts.
\paragraph{Annotation Instruction}
Here we present the human evaluation standard:

\noindent \textbf{Fluency(Flue.)}: Whether the generated news is fluent and easy to understand.
\begin{enumerate}
    \item The system’s result does not make sense and it is unreadable.
    \item Choose this score when you are hesitant between score 1 and score 3.
    \item The system’s result contains minor errors but they do not affect your understanding.
    \item Choose this score when you are hesitant between score 3 and score 5.
    \item The system’s result is human-like, grammatically correct, and very easy to understand. 
\end{enumerate}

\noindent \textbf{Relevance (Relv.):} Whether the generated news is related to the given image and the corresponding title.
\begin{enumerate}
    \item The system’s result is completely irrelevant to the given image.
    \item Choose this score when you are hesitant between score 1 and score 3. 
    \item The system’s result is partially related to the image and some of its content can be found in the image. 
    \item Choose this score when you are hesitant between score 3 and score 5. 
    \item The system’s result is very related to the given image and contains a diverse set of concepts in the image. 
\end{enumerate}

\noindent \textbf{Sentiment (Sent.):} Does the generated news have positive or negative sentiment.
\begin{itemize}
    \item \textbf{Positive:} The system’s result has a positive sentiment.
    \item \textbf{Negative:} The system’s result has a negative sentiment.
    \item \textbf{Can't Tell:} The system’s result is neither negative nor positive.
\end{itemize}

\end{document}